\pdfoutput=1

\documentclass[11pt]{article}

\usepackage[]{acl}

\usepackage[tikz]{bclogo}
\usepackage{times}
\usepackage{latexsym}

\usepackage[T1]{fontenc}

\usepackage[utf8]{inputenc}

\usepackage{microtype}
\usepackage{multirow}
\usepackage{booktabs}
\usepackage{amsmath}
\usepackage{enumerate}
\usepackage{enumitem}
\usepackage{pdfpages}
\usepackage{newtxmath}

\usepackage{inconsolata}
\usepackage{makecell}
\usepackage{longtable}
\usepackage{amsmath,amsfonts,bm}

\newcommand{\ie}{{\em i.e.,}}
\newcommand{\eg}{{\em e.g.,}}
\newcommand{\wrt}{\emph{w.r.t.}}

\newcommand{\Ni}{({\em i})~}
\newcommand{\Nii}{({\em ii})~}
\newcommand{\Niii}{({\em iii})~}

\makeatletter   
\newcommand{\sveryshortarrow}[1][3pt]{\mathrel{%
    \vcenter{\hbox{\rule[-.5\fontdimen8\scriptfont3]
               {\scriptratio\dimexpr#1\relax}{\fontdimen8\scriptfont3}}}%
   \mkern-4mu\hbox{\let\f@size\sf@size\usefont{U}{lasy}{m}{n}\symbol{41}}}}
\makeatother

\def\eqref#1{equation~\ref{#1}}

\def\1{\bm{1}}

\def\m1{{\bm{1}}}

\DeclareMathAlphabet{\mathsfit}{\encodingdefault}{\sfdefault}{m}{sl}
\SetMathAlphabet{\mathsfit}{bold}{\encodingdefault}{\sfdefault}{bx}{n}

\def\gL{{\mathcal{L}}}

\usepackage[nameinlink]{cleveref}
\crefformat{section}{\S#2#1#3} 
\crefname{algorithm}{Alg.}{Algs.}
\crefformat{subsection}{\S#2#1#3}
\Crefname{equation}{Eq.}{Eqs.}
\Crefname{figure}{Fig.}{Figs.}
\title{Learning to Initialize: Can Meta Learning Improve \\ 
Cross-task Generalization in Prompt Tuning?}

\author{Chengwei Qin$^\clubsuit$, Qian Li$^\vardiamondsuit$, Ruochen Zhao$^\clubsuit$, Shafiq Joty$^\clubsuit$$^\spadesuit$\\
$^\clubsuit$ Nanyang Technological University\\
$^\spadesuit$ Salesforce AI \\
$^\vardiamondsuit$ Northeastern University \\
\texttt{\{chengwei003@e.ntu, ruochen002@e.ntu, srjoty@ntu\}.edu.sg}\\ \texttt{qianli@stumail.neu.edu.cn}
}

\definecolor{msftBlack}{RGB}{0,0,0}
\newcommand{\finding}[1]{
	\begin{bclogo}[couleur= msftBlack!10, epBord=1, arrondi=0.1, logo=\bclampe, marge=2, ombre=true, blur, couleurBord=msftBlack!20, tailleOndu=3, sousTitre={\em #1}]{} 
	\end{bclogo}
}

\begin{document}
\maketitle
\begin{abstract}

\emph{Prompt tuning} (PT) which only tunes the embeddings of an additional sequence of tokens per task, keeping the pre-trained language model (PLM) frozen, has shown remarkable performance in {few-shot learning}. Despite this, PT has been shown to rely heavily on good initialization of the prompt embeddings. In this work, we study \emph{meta prompt tuning} (MPT) to systematically explore how {meta-learning} can help improve (if it can) cross-task generalization in PT through learning to initialize the prompt embeddings from other relevant tasks. We empirically analyze a representative set of meta learning algorithms in a wide range of adaptation settings with different source/target task configurations on a large set of few-shot tasks. With extensive experiments and analysis, we demonstrate the effectiveness of MPT. We find the improvement to be significant particularly on  classification tasks. For other kinds of tasks such as question answering, we observe that while MPT can outperform PT in most cases, it does not always outperform multi-task learning. {We further provide an in-depth analysis from the perspective of task similarity.}

\end{abstract}

\section{Introduction} \label{sec:intro}

Humans can easily learn to perform new tasks with only few data by leveraging previously acquired knowledge from other relevant tasks. Such capability is a hallmark of human intelligence \citep{Carey78}. However, when it comes to the models, they often face over-fitting issues when they are tasked to learn from a few labeled examples \citep{lake_ullman_tenenbaum_gershman_2017,linzen-2020-accelerate}, a problem commonly termed as \emph{few-shot learning} (FSL).

With the recent advancements in developing large-scale pre-trained  language models (PLMs), prompt-based methods have shown promising results in FSL. \citet{brown2020language} show that by virtue of in-context (meta) learning, a frozen GPT-3 model can achieve good results on a variety of few-shot tasks through manually designed \emph{prompts}, which are task instructions along with a few examples expressed in natural language. However, the performance of in-context learning has been shown to be highly sensitive to the design of such ``discrete'' prompts \citep{pmlr-v139-zhao21c}. It is also limited by the maximum sequence length supported by the PLMs \cite{li-liang-2021-prefix}. Down this line, efforts have been made on automatically searching and optimizing for discrete prompts \citep{shin-etal-2020-autoprompt,schick-schutze-2021-exploiting,gao-etal-2021-making}. 

\begin{figure}[t]
    \centering
    \includegraphics[width=0.48\textwidth]{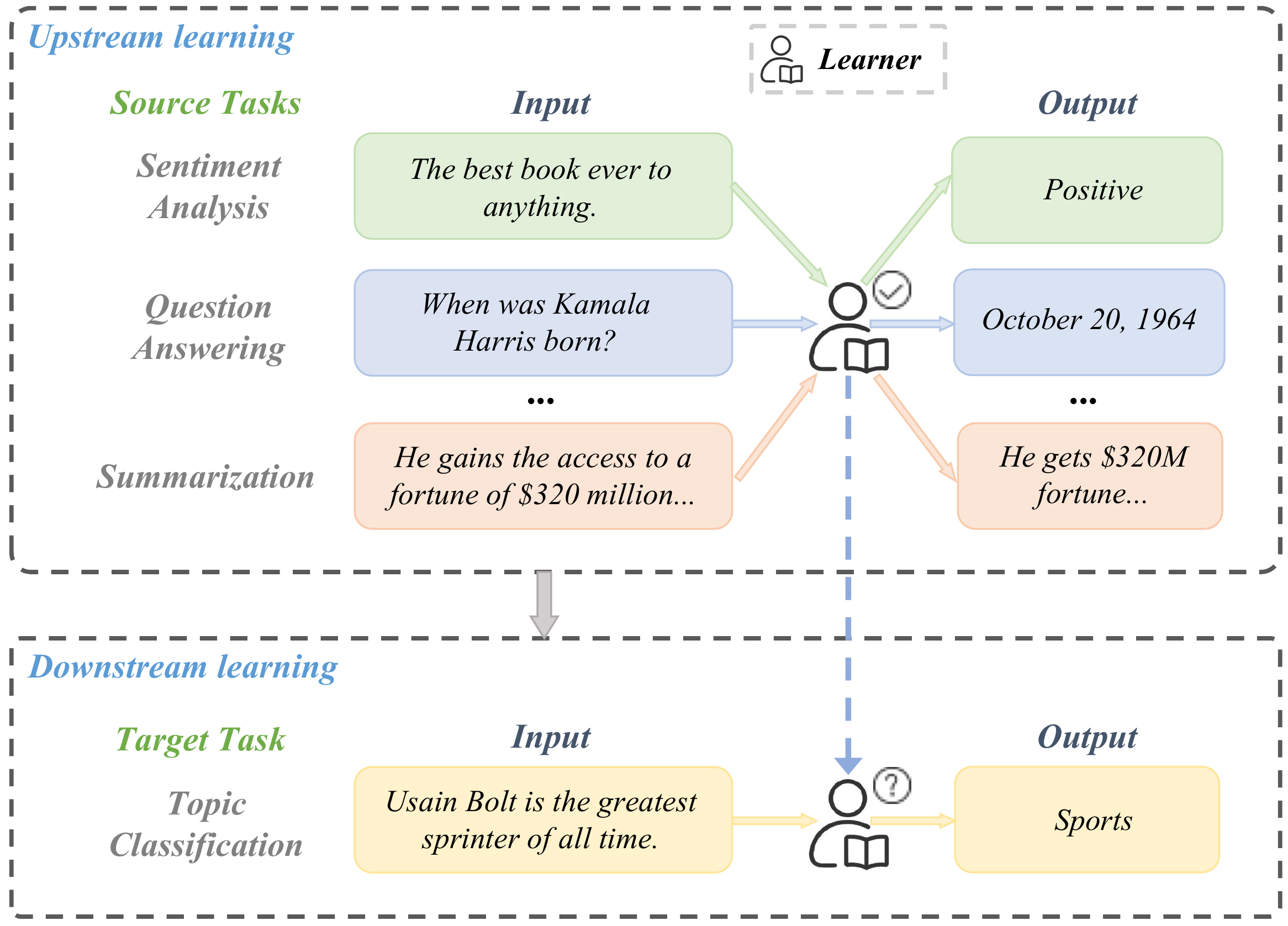}
    \caption{ Illustration of cross-task generalization, where the model is expected to learn an unseen \emph{target} task given the knowledge acquired from previously learned \emph{source} tasks.}
    \label{fig:ctg}
\end{figure}

As an alternative to discrete prompts, recent efforts attempt to learn ``soft'' prompts that add additional trainable parameters \citep{liu2021gpt,li-liang-2021-prefix,lester-etal-2021-power}, showing better results  than discrete prompts \citep{liu2021pre}. \citet{lester-etal-2021-power} introduce \emph{prompt tuning} (PT) that prepends a sequence of \emph{tunable} tokens to the input and optimize their embeddings keeping the PLM frozen. Despite its strong few-shot performance, PT has been shown to be sensitive to the initialization of the embeddings, which might limit its practical application  \citep{qin2022lfpt}. To address this, \citet{gu-etal-2022-ppt} propose \emph{pre-trained prompt tuning} (PPT) to pre-train soft prompts using self-supervised tasks on unlabeled data. It relies on carefully designed pre-training tasks tailored to the downstream tasks, and the pre-training objectives are only applicable to classification tasks.   
\citet{vu-etal-2022-spot} introduce \emph{soft prompt transfer} (SPoT), which uses the soft prompts learned from a set of source tasks through multi-task learning to initialize the prompt for a target task. Both PPT and SPoT demonstrate \emph{cross-task generalization} (\Cref{fig:ctg}) -- learning of a new task can benefit from learning of other related tasks \citep{ye-etal-2021-crossfit}.


In a recent survey, \citet{lee2022meta} claim that \emph{meta learning} \cite{schmidhuber:1987:srl} can play an important role for cross-task generalization in NLP.\footnote{Unless otherwise specified, by meta learning in this paper we generally refer to the optimization-based meta learning algorithms, and use more specific names for the other kinds such as \emph{in-context learning} for black-box meta learning and \emph{metric learning} for non-parametric meta learning.} Different from multi-task learning which considers the performance on the source tasks to learn the initial parameters, meta learning aims to find initial parameters suitable for adapting to a target few-shot task. Hence, it could outperform multi-task learning in several scenarios with \emph{full-model} finetuning  \citep{dou-etal-2019-investigating,chen-etal-2020-low}. However, to our knowledge, there is no systematic study on the role of meta learning on PT. In a {recent} work, \citet{huang2022learning} adopt MAML \citep{finn2017model}  for pre-training soft prompts. One major limitation of their study is that it is limited to only one type of meta learning algorithm and only sentiment classification tasks, lacking comprehensive understanding of cross-task generalization.{ \citet{min-etal-2022-metaicl} and \citet{chen-etal-2022-meta} show the effectiveness of in-context learning for PLMs, whereas we mainly focus on optimization-based meta learning.}

To systematically study meta prompt tuning (MPT) for cross-task generalization, we conduct experiments on a large collection of few-shot tasks involving different types of datasets with a unified text-to-text format \citep{ye-etal-2021-crossfit}. We investigate a wide range of adaptation settings with different source/target task types, which helps better understand the capability and limitation of meta learning in PT. With extensive experiments, we aim to address the following research questions:
\begin{itemize}[leftmargin=*,topsep=2pt,itemsep=2pt,parsep=0pt]
    \item \textbf{Q1.} Can MPT improve cross-task generalization in PT? Is it better than multi-task learning?
    \item \textbf{Q2.} What happens with more labelled data for source/target tasks (beyond few-shot settings)?
    \item \textbf{Q3.} Does it help with more diverse source tasks?
    \item \textbf{Q4.} Is the performance gain of MPT consistent across different backbone models?
\end{itemize}

To answer these questions, we empirically analyze MAML \citep{finn2017model}, FoMAML and Reptile \citep{nichol2018first}, which constitute a representative set of meta learning methods. Experimental results show that MPT can indeed help cross-task generalization, \eg\ MAML improves the performance of PT by more than 20$\%$ on classification tasks. However, we also notice that MPT does not always outperform multi-task learning, especially on non-classification tasks. {We provide an in-depth analysis from the perspective of task similarity.} 
{As for Q2, we find that MPT does benefit cross-task generalization beyond few-shot settings. For Q3, we observe that increasing the diversity of source tasks does not necessarily improve cross-task generalization. Finally, the consistent gain of MPT across different models shows its robustness to model {type and} size.} 
In summary, the two main contributions of this work are:
\begin{itemize}[leftmargin=*,topsep=2pt,itemsep=2pt,parsep=0pt]
    \item To the best of our knowledge, we are the first to extensively explore how meta learning helps cross-task generalization in prompt tuning.
    \item With extensive experiments and analysis, we show the effectiveness and limitation of meta prompt tuning in various source/target settings. 
\end{itemize}
\section{Related Work}

\paragraph{Few-shot Learning (FSL)} FSL aims to learn a task with only a few labeled examples, which often leads to the over-fitting problem. Existing methods to address this problem mainly focus on optimizing the hypothesis space of the few-shot tasks \citep{triantafillou2017few,finn2017model,hu-etal-2018-shot} or augmenting the few-shot data \citep{gao2020neural,qin-joty-2022-continual}. Recently, large-scale pre-trained language models (PLMs) have demonstrated strong FSL ability through prompt-based methods, including both discrete  \citep{brown2020language} and soft prompts \citep{lester-etal-2021-power}.

\paragraph{Prompt-based Learning (PL)} PL is a new paradigm which prepends a task-specific template or prompt to the input for learning new tasks \citep{liu2021pre}. Initial PL methods mainly focus on designing, searching or optimizing discrete prompts \citep{brown2020language,shin-etal-2020-autoprompt,gao-etal-2021-making}. However, discrete prompts are hard to optimize. To solve this, recent PL methods attempt to optimize prompts in a continuous space, \ie\ learn soft prompts \citep{li-liang-2021-prefix,liu2021gpt,lester-etal-2021-power}, showing impressive FSL performance \citep{qin2022lfpt}. In addition to prompt design, several recent studies have explored the applications \citep{
zhu2022prompt,li-etal-2022-learning-transfer,ding-etal-2023-gpt,qin2023chatgpt,zhao2023verify} and analysis \citep{zhong-etal-2021-adapting-language,le-scao-rush-2021-many} of PL.

\paragraph{Meta Learning} 

Meta Learning or learning to learn, has been applied to boost few-shot performance on various NLP tasks, \eg\ relation extraction \citep{han-etal-2018-fewrel} and machine translation \citep{gu-etal-2018-meta}. Meta learning algorithms can be 
divided into three main categories. First, \emph{black-box} methods adopt additional meta learners to help adaptation \citep{santoro2016meta,garnelo2018conditional,mishra2018a,brown2020language}. Second, \emph{non-parametric} methods explore how to learn metrics that can compare the distances between different samples, \ie\ learning to compare \citep{koch2015siamese,vinyals2016matching,snell2017prototypical}. Finally, \emph{optimization-based} methods aim to learn better parameter initialization to effectively and efficiently adapt to unseen tasks, \ie\ learning to initialize \citep{finn2017model,nichol2018first,kedia-etal-2021-beyond-reptile}. \citet{lee2022meta} claim that meta learning can be effective for cross-task generalization, especially the optimization-based methods. They can be applied to various problems in a model-agnostic way to improve FSL on target tasks with model fine-tuning \citep{ye-etal-2021-crossfit}.

\paragraph{Summary.} Existing work shows that meta learning can improve cross-task few-shot generalization with full model fine-tuning. However, there is no systematic study on whether (and how) meta learning can do so with prompt tuning of PLMs. To fill this research gap, our work provides a comprehensive understanding of the effectiveness and limitation of meta learning in prompt tuning.

\section{Preliminaries}

In this section, we revisit the basics about prompt tuning and optimization-based meta learning.

\subsection{Prompt Tuning}

Following \citet{lester-etal-2021-power}, we reframe all tasks into a text-to-text format. Given a training dataset $\mathcal{D}^{tr} = \{(X_1,Y_1),...,(X_n,Y_n)\}$ for a task $\mathcal{T}$, different from traditional model fine-tuning, prompt tuning (PT) is a parameter-efficient learning method which freezes the PLM $\theta$ and prepends the input text $X_i$ with a sequence of \emph{tunable} soft tokens $P$, parameterized by prompt embeddings $\phi$. The prompt embeddings $\phi$ are initialized from the vocabulary of the PLM and optimized through gradient descent with the following objective:
\begin{equation}
\small 
\gL_{\phi}^{\mathcal{T}} = \mathcal{L}(\phi, \mathcal{D}^{tr}) =  - \sum_{i=1}^{n} \log p (Y_i|[P,X_i], \phi, \theta) \label{eq:pt}
\end{equation}

\subsection{Optimization-based Meta Learning}

The main goal of optimization-based meta learning (or learning to initialize), is to learn better initial parameters that can effectively and efficiently adapt to a new task $\mathcal{T}^\text{new}$ with limited data. We denote the initial parameters (meta-parameters) as $\phi^*$. 

To obtain $\phi^*$, the model needs to learn from a series of  \emph{meta-training} tasks $\mathcal{T}^{\text{meta}} = \{\mathcal{T}_1,...,\mathcal{T}_n\}$. The dataset $\mathcal{D}_i$ of  each task $\mathcal{T}_i$ is divided into two disjoint sets: a \emph{support set} $\mathcal{S}_i$ and a \emph{query set} $\mathcal{Q}_i$. The objective for learning $\phi^*$ is 
\begin{equation}
\small 
\phi^* = \mathop{\arg\min}\limits_{\phi} \sum_{\mathcal{T}_i \in \mathcal{T}^{\text{meta}}} \gL \Big( \underbrace{\phi - \alpha \nabla_{\phi} \mathcal{L}(\phi, \mathcal{S}_i)}_{\text{inner update}}, \mathcal{Q}_i \Big) \label{eq:meta} 
\end{equation}
where $\mathcal{L}$ is the objective function defined in \Cref{eq:pt}, $\phi$ is the set of parameters to meta-learn and $\alpha$ is the inner learning rate. Denoting the overall loss as $\gL_{\phi}^{\mathcal{T}^{\text{meta}}} = \sum_{\mathcal{T}_i \in \mathcal{T}^{\text{meta}}} \gL (\phi', \mathcal{Q}_i)$ with $\phi'$ being the inner-updated value of $\phi$, we use gradient descent to update $\phi$ further in the meta-training stage:
\begin{equation}
\phi = \phi - \beta \nabla_{\phi} \gL_{\phi}^{\mathcal{T}^{\text{meta}}} \label{eq:obj} 
\end{equation} 
where $\beta$ is the outer learning rate. This is actually the Model-Agnostic Meta-Learning or MAML \citep{finn2017model}. Notice that optimizing \Cref{eq:obj} requires calculating second-order gradients, which can be quite memory-consuming. To alleviate this, First-order MAML (FoMAML) and Reptile \citep{nichol2018first} are proposed to use first-order approximations, allowing lower memory costs. 

After the meta-training stage, $\phi^*$ serves as the initial parameters for learning an unseen \emph{meta-testing} task $\mathcal{T}^\text{new}$ which is usually few-shot. 

\section{Approach}

In this section, we first introduce the problem setting and evaluation metric. Then, we illustrate the key methods for meta prompt tuning (MPT).

\subsection{Problem Setting}
To evaluate cross-task generalization in prompt tuning, we select a large and diverse collection of few-shot tasks from \citet{ye-etal-2021-crossfit}, covering various types including classification, question answering and generation. We partition the set of all tasks $\mathcal{T}^\text{all}$ into two disjoint parts: source tasks $\mathcal{T}^\text{src}$ and target tasks $\mathcal{T}^\text{tgt}$.  Details of the tasks and partitions are provided later in our experiment setup (\cref{sec:exp_set}).

Following \citet{min-etal-2022-metaicl}, we can divide the whole learning process into two stages (\Cref{fig:ctg}):
\paragraph{$\bullet$ Upstream learning on source tasks} In this stage, the model has access to $\mathcal{T}^\text{src}$, which is regarded as \emph{meta-training} tasks $\mathcal{T}^{\text{meta}}$ in \Cref{eq:meta}. We divide the dataset $\mathcal{D}_i$ of every source task $\mathcal{T}_i$ into training (or support) and validation (or query) sets, and conduct optimization-based meta learning or multi-task learning on these sets to obtain meta-parameters $\phi^*$. Note that we use both support and query sets for model training in multi-task learning to ensure fair data access for both methods.

\paragraph{$\bullet$ Downstream learning on target tasks} After the upstream learning stage, we use the learned meta-parameters $\phi^*$ as the initial point for learning target tasks  $\mathcal{T}^\text{tgt}$. Every target task $\mathcal{T}_k$ has its own training set $\mathcal{D}^{\text{tr}}_k$, validation set $\mathcal{D}^{\text{val}}_k$, and test set $\mathcal{D}^{\text{test}}_k$. The model is required to learn from $\mathcal{D}^{\text{tr}}_k$ via prompt tuning and will be evaluated on $\mathcal{D}^{\text{test}}_k$. The performance on $\mathcal{D}^{\text{val}}_k$ is used for hyper-parameters tuning and model selection. 

This two-stage learning paradigm can naturally reflect cross-task generalization where the model needs to learn an unseen task given previously acquired knowledge from other tasks.

\subsection{Evaluation Metric}

We evaluate the model performance on a set of target tasks $\mathcal{T}^\text{tgt}$. As $\mathcal{T}^\text{tgt}$ may cover various task types, simply averaging the performance of different target tasks is unreasonable. Following \citet{ye-etal-2021-crossfit}, we use \emph{average relative gain} (ARG) as the main evaluation metric. We first calculate \emph{relative gain} (RG) for each target task, \ie\ relative performance improvement before and after applying the upstream (meta or multi-task) learning  on the source tasks. Then we average the relative gains of all target tasks to obtain the final result which indicates the overall performance improvement.

\subsection{Meta Prompt Tuning (MPT)}

\begin{figure}[t]
    \centering
    \includegraphics[width=0.48\textwidth]{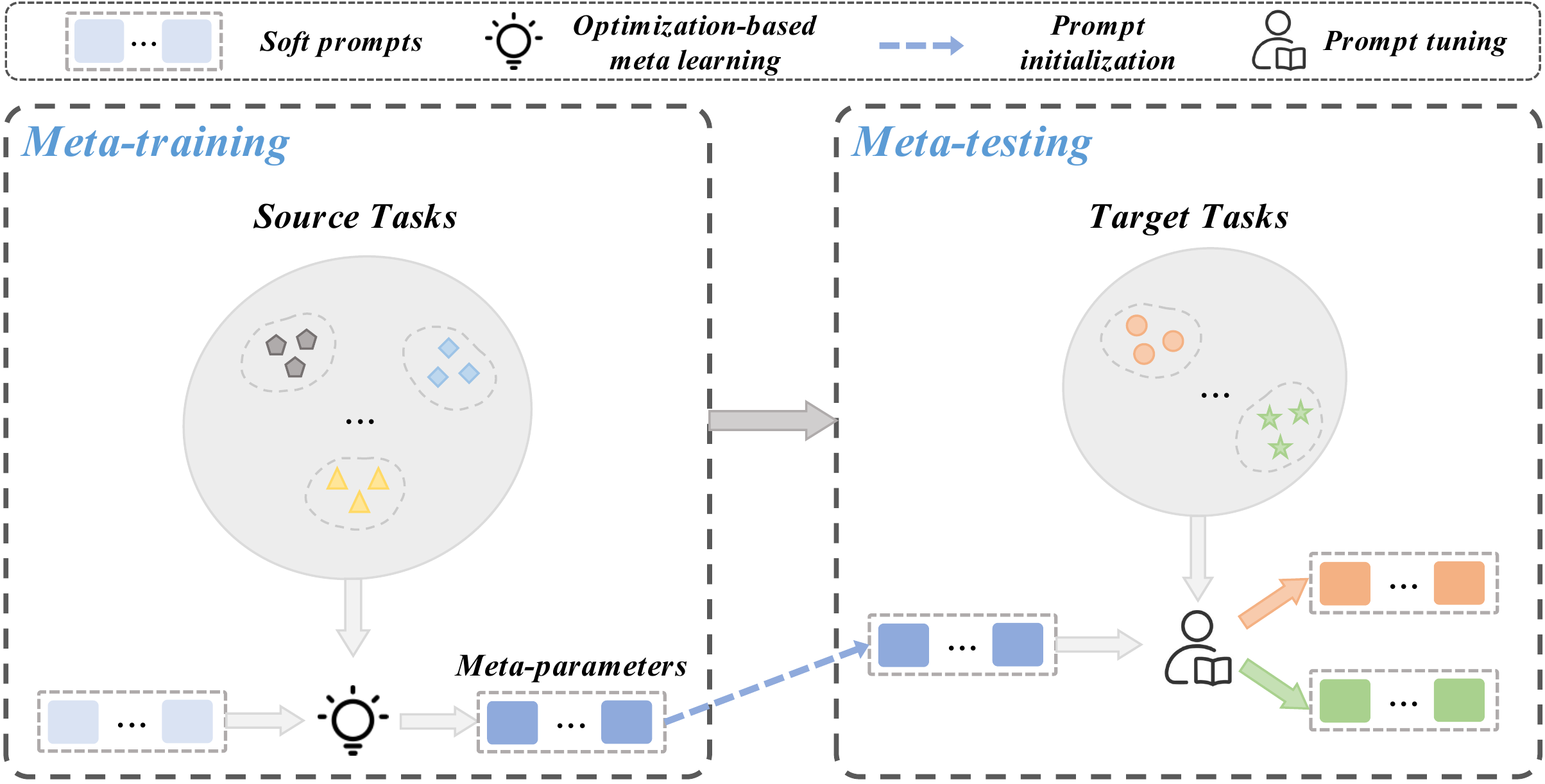}
    \caption{Overview of Meta Prompt Tuning (MPT). In the meta-training stage, we conduct optimization-based meta learning on source tasks to obtain meta-parameters (\ie\ soft prompts). The meta-parameters will then be used to initialize prompt embeddings for learning unseen target tasks in the meta-testing stage.}
    \label{fig:metapt}
\end{figure}

As shown in \Cref{fig:metapt}, the key idea of MPT is to apply optimization-based meta-training as {upstream learning} to a set of source tasks  in order to learn meta parameters, which in this case are prompt embeddings. The learned prompt embeddings serve as the initialization for learning unseen target tasks, referred to as meta-testing or downstream learning.

\subsubsection{Meta-training}
We meta-train the prompt embeddings on source tasks $\mathcal{T}^\text{src}$. Without loss of generality, we take MAML \citep{finn2017model} as an example. For every iteration, we first sample one source task $\mathcal{T}_i$ which has a support set $\mathcal{S}_i$ and a query set $\mathcal{Q}_i$. Then we sample a support batch $\mathcal{B}_s$ from $\mathcal{S}_i$ and a query batch  $\mathcal{B}_q$ from $\mathcal{Q}_i$. Denoting the trainable prompt embeddings as $\phi$, $\mathcal{B}_s$ and $\mathcal{B}_q$ are used for one gradient update with the following objective:
\begin{equation}
\begin{split}
\gL_{\phi}^i  &= \gL (\phi - \alpha \nabla_{\phi} \mathcal{L}(\phi, \mathcal{B}_s), \mathcal{B}_q) \\
\phi &= \phi - \beta \nabla_{\phi} \gL_{\phi}^i 
\end{split}
\end{equation} 
where $\mathcal{L}$ is the task loss defined in \Cref{eq:pt}, and $\alpha$ and $\beta$ are inner and outer learning rates, respectively. 
During the meta-training stage, we iterate over tasks in $\mathcal{T}^\text{src}$ to update prompt embeddings $\phi$ for a fixed number of steps. The learned meta-parameters $\phi^*$ is used in the meta-testing stage.

\subsubsection{Meta-testing}
In meta-testing, the model is expected to learn unseen target tasks $\mathcal{T}^\text{tgt}$. For each target task $\mathcal{T}_k$, we use the learned meta-parameters $\phi^*$ to initialize the prompt embeddings for the task. Denoting the training set of $\mathcal{T}_k$ as $\mathcal{D}^{\text{tr}}_k$, the learning objective during meta testing is defined as:
\begin{equation}
\mathcal{L}_{\phi^*}(\mathcal{D}^{\text{tr}}_k) =  - \sum_{i=1}^{n} \log p (Y_i|[P^*,X_i], \phi^*, \theta)
\end{equation}
where $\theta$ is the frozen PLM, $(X_i,Y_i) \sim \mathcal{D}^{\text{tr}}_k$ is a training sample and $P^*$ are the prompt tokens. 

We evaluate the model with the best validation performance on the test set and calculate average relative gain on the 
test sets of $\mathcal{T}^\text{tgt}$.
\section{Experimental Setup} \label{sec:exp_set}
We first describe the source/target task partitions, and then introduce methods compared in our work. Finally, we present the implementation details.

\subsection{Task Partitions}  \label{sec:partition}

We experiment with ten different source/target task partitions as shown in \Cref{partition}. Depending on the type of the target tasks, we can divide these ten settings into several groups:

\begin{table}[t]
    \centering \footnotesize
    \resizebox{1.0\linewidth}{!}{
    \begin{tabular}{lrlr}
        \toprule
            \multicolumn{2}{c}{\textbf{Source}} & \multicolumn{2}{c}{\textbf{Target}} \\
            \cmidrule(lr){1-2} \cmidrule(lr){3-4}
            Setting & \#tasks  & Setting & \#tasks\\
        \midrule
            Random & 114 & Random & 20 \\
        \cmidrule(lr){1-4}
            Classification (Cls)   & 45  & \multirow{3}{*}{Classification} & \multirow{3}{*}{10} \\
            Both (Cls + Non-Cls) & 23 + 22 & & \\
            Non-Classification & 45 & & \\
            \cmidrule(lr){1-4}
            Classification & 45  & \multirow{3}{*}{Non-Classification} & \multirow{3}{*}{12} \\
            Both (Cls + Non-Cls) & 23 + 22 & & \\
            Non-Classification & 45 & & \\
        \cmidrule(lr){1-4}
            QA & 22 & \multirow{2}{*}{QA} & \multirow{2}{*}{15} \\
            Non-QA & 33  & & \\
        \cmidrule(lr){1-4}
            Non-Paraphrase Cls & 60  & Paraphrase & 4 \\
        \bottomrule
    \end{tabular}
    }
    \caption{Statistics of ten distinct source/target task partitions. Appendix~\ref{sec:full-data-list} for details about each partition.} 
    \label{partition}
\end{table}

\begin{itemize}[leftmargin=*,topsep=3pt,itemsep=3pt,parsep=0pt]
\item \textbf{R$\rightarrow$R (Random$\rightarrow$Random)}: We first experiment with the R$\rightarrow$R setting where both source and target tasks are randomly selected, meaning that they can cover any task type. This setting mimics the learning paradigm of humans and reflects whether cross-task generalization can help obtain a general-purpose few-shot learner.
\item \textbf{X$\rightarrow$Cls (X={Cls, Both, Non-Cls})}: The target tasks involve classification, while the source tasks can be classification, non-classification tasks or both. This setting helps us better understand the influence of the source task distribution. 
\item \textbf{X$\rightarrow$Non-Cls (X={Cls, Both, Non-Cls})}: The only difference between this and the previous setting is the type of target tasks. We investigate how meta learning improves cross-task generalization when target tasks are non-classification tasks.
\item \textbf{X$\rightarrow$QA (X={QA, Non-QA})}: Compared to the previous one, this group is more fine-grained. We only select target tasks from question answering (QA) instead of all non-classification tasks. We conduct experiment on different source task types, including QA and Non-QA tasks.
\item \textbf{NP$\rightarrow$P (Non-Paraphrase Cls$\rightarrow$Paraphrase)}: This group has the  finest granularity in our setting. We choose paraphrase identification which is a sub-category of classification as the target, and non-paraphrase classification as the source. The final two groups help understand how meta learning performs in more fine-grained scenarios. 
\end{itemize}

Note that we ensure that there is no overlap between the source and target tasks. Following \citet{ye-etal-2021-crossfit}, we use 16 samples per class in the training (or support) and validation (or query) sets for classification tasks, and 32 samples per set for non-classification tasks. For every task, we sample the training and validation sets 5 times with different random seeds to reduce variance in few-shot evaluation and cover more diverse samples in upstream learning. We provide full details of tasks and partitions in Appendix~\ref{sec:full-data-list}.

\subsection{Methods Compared}
We mainly use T5-Large \citep{raffel2019exploring} as the backbone language model and compare the following methods in our work.

\begin{itemize}[leftmargin=*,topsep=4pt,itemsep=4pt,parsep=0pt]
\item \textbf{Prompt Tuning (PT) on target tasks.} It is our baseline without the upstream learning. We directly apply PT \citep{lester-etal-2021-power} to target tasks and use its performance as the basis for computing average relative gain for other methods.
\item \textbf{Model-Agnostic Meta-Learning (MAML).} We apply MAML \citep{finn2017model} in the upstream learning (meta-training) stage. The learned meta-parameters are used to initialize prompt embeddings for learning target tasks.
\item \textbf{First-order MAML (FoMAML) and Reptile.} We also investigate two first-order meta learning algorithms:  FoMAML \citep{finn2017model} and Reptile \citep{nichol2018first}. Compared to MAML, they are more memory-efficient.
\item \textbf{Multi-task learning (MTL).} We conduct multi-task learning on source tasks instead of meta learning to obtain initial parameters. This is a straight-forward yet effective method as demonstrated by \citet{vu-etal-2022-spot}.
\item \textbf{Fine-tuning on target tasks.} Fine-tuning is the dominant paradigm where the whole language model is tuned for learning target tasks. We include it to verify whether cross-task generalization can help PT outperform fine-tuning.
\end{itemize}

In addition, we conduct experiments with {different backbone models to verify MPT's robustness.}


\subsection{Implementation Details} \label{sec:hyper}
All our methods are implemented with PyTorch/Transformers library \citep{wolf-etal-2020-transformers}. We use higher library \citep{grefenstette2019generalized} for higher-order optimization in meta learning methods. The prompt length in PT is set to 100 tokens following \citet{lester-etal-2021-power}. For meta-training, we set the inner and outer learning rates to $3\mathrm{e}{-5}$ and $5\mathrm{e}{-1}$, respectively. We use 5000 for total training steps. We set the inner batch size to 2, 4 and 4, and inner update steps to 1, 1 and 10 for MAML, FoMAML and Reptile, respectively. For multi-task learning, we set the learning rate, batch size and number of epochs to $5\mathrm{e}{-1}$, $4$ and $20$, respectively. For MAML, we select the inner learning rate from $\{2\mathrm{e}{-5},3\mathrm{e}{-5},5\mathrm{e}{-5}\}$, the outer learning rate from $\{2\mathrm{e}{-1},3\mathrm{e}{-1},5\mathrm{e}{-1}\}$, and total training steps from $\{2500,5000,10000\}$. We adopt the same three hyperparameters for FoMAML and Reptile. The search range for the inner update steps of Reptile is $\{2,4,6,8,10\}$. For multi-task learning, we select the learning rate from $\{2\mathrm{e}{-1},3\mathrm{e}{-1},5\mathrm{e}{-1}\}$, the batch size from $\{2,4,6,8\}$, and the number of epochs from $\{5,10,20\}$.

For downstream learning, we mainly follow the settings in \citet{ye-etal-2021-crossfit}. For prompt tuning, we select the learning rate from $\{5\mathrm{e}{-1},4\mathrm{e}{-1},3\mathrm{e}{-1},2\mathrm{e}{-1}\}$ based on the validation performance. For fine-tuning, the search range for the learning rate is $\{5\mathrm{e}{-4},3\mathrm{e}{-4},2\mathrm{e}{-4},1\mathrm{e}{-4}\}$. We set the batch size, total training steps and evaluation interval to 8, 3000 and 50, respectively.

Since it is infeasible to search for optimal hyperparameters for each of the meta- and multi-task learning  methods in each of the settings, we select them based on the R$\rightarrow$R setting. We randomly select 5 tasks that are not in the source and target sets as validation tasks for hyperparameter search. The hyperparameters with best validation performance (ARG) are used for upstream learning. We select the inner learning rate, the outer learning rate and total training steps for MAML and adopt the same three hyperparameters for FoMAML and Reptile.

\begin{table*}[t]
    \centering
    \resizebox{1.0\linewidth}{!}{
    \begin{tabular}{
        l @{\hspace{2em}}
        cccccccccc
        }
        \toprule
            \textbf{Method} &
            \makecell[c]{R$\rightarrow$R} &
            \makecell[c]{Cls \\ $\rightarrow$Cls} & \makecell[c]{Both \\ $\rightarrow$Cls} & \makecell[c]{Non-Cls \\ $\rightarrow$Cls} &
            \makecell[c]{Cls \\ $\rightarrow$Non-Cls} & \makecell[c]{Both \\ $\rightarrow$Non-Cls} & \makecell[c]{Non-Cls \\ $\rightarrow$Non-Cls} &
            \makecell[c]{QA \\ $\rightarrow$QA} &
            \makecell[c]{Non-QA \\ $\rightarrow$QA} &
            \makecell[c]{NP \\ $\rightarrow$P} \\
        \midrule
            MAML & $\mathbf{8.78}_{\pm 0.69}$ & $\mathbf{20.16}_{\pm 0.84}$ & $10.57_{\pm 1.03}$ & $6.34_{\pm 0.48}$ & $0.32_{\pm 0.04}$ & $7.54_{\pm 0.73}$ & $6.71_{\pm 0.39}$ & $-16.59_{\pm 1.36}$ & $3.26_{\pm 0.24}$ & $11.14_{\pm 0.93}$ \\
            FoMAML & $1.24_{\pm 0.18}$ & $18.80_{\pm 1.13}$ & $\mathbf{17.84}_{\pm 1.21}$ & $7.32_{\pm 0.42}$ & $6.42_{\pm 0.51}$ & $9.81_{\pm 0.64}$ & $3.88_{\pm 0.31}$ & $16.63_{\pm 1.58}$ & $9.83_{\pm 0.76}$ & $-0.68_{\pm 0.07}$ \\
            Reptile & $8.42_{\pm 0.46}$ & $-5.17_{\pm 0.71}$ & $-4.18_{\pm 0.37}$ & $2.42_{\pm 0.21}$ & $-1.54_{\pm 0.18}$ & $-3.38_{\pm 0.49}$ & $0.78_{\pm 0.07}$ & $0.77_{\pm 0.09}$ & $-0.09_{\pm 0.01}$ & $\mathbf{20.44}_{\pm 1.34}$ \\
        \cmidrule{1-11}
            Multi-task learning & $7.14_{\pm 0.62}$ & $-5.64_{\pm 0.92}$ & $5.73_{\pm 0.43}$ & $4.97_{\pm 0.39}$ & $\mathbf{8.51}_{\pm 1.16}$ & $\mathbf{13.47}_{\pm 0.97}$ & $\mathbf{19.67}_{\pm 1.72}$ & $\mathbf{25.65}_{\pm 1.93}$ & $\mathbf{17.23}_{\pm 1.08}$ & $-5.19_{\pm 0.86}$ \\
        \cmidrule{1-11}
            Fine-tuning          & $-12.61_{\pm 1.57}$ & $16.02_{\pm 1.44}$ & $16.02_{\pm 1.44}$ & $\mathbf{16.02}_{\pm 1.44}$ & $-35.70_{\pm 2.73}$ & $-35.70_{\pm 2.73}$ & $-35.70_{\pm 2.73}$ & $-47.37_{\pm 2.97}$ & $-47.37_{\pm 2.97}$ & $1.56_{\pm 0.12}$ \\
        \bottomrule
    \end{tabular}
    }
    \caption{ \textbf{Average relative gain (ARG $\%$) of different methods with respect to prompt tuning (PT) in various settings.} \textbf{Bold} indicates the best ARG score. `Cls', `QA', `P' and `NP' respectively stand for `classification', `question answering',  `paraphrase' and `non-paraphrase classification'.
    }\label{tab:main-result}
\end{table*}

\section{Results and Analysis} \label{sec:exp_res}

We now address the four research questions asked before in \Cref{sec:intro} with empirical results. 

\finding{\small \textbf{Q1.} Can meta prompt tuning improve cross-task generalization? Is it better than multi-task learning? \label{Q1}
}

The ARG of different methods \wrt\ PT in various settings are shown in \Cref{tab:main-result}; more detailed results on every target task are in Appendix~\ref{sec:detail}.

\paragraph{$\bullet$ MPT can indeed help cross-task generalization.} From the results in \Cref{tab:main-result}, we observe that MPT outperforms the baseline PT in most cases with +ve ARG scores. Out of 30 different runs for three meta learning methods in ten different settings (see the 1st block of results), MPT achieves better performance than PT in 23 runs, demonstrating its effectiveness in cross-task generalization. 

For the R$\rightarrow$R setting, MAML achieves the best performance, showing that it is a good general-purpose few-shot learner. For adapting to classification tasks, MAML outperforms PT by \textbf{20.16}$\%$ if the prompt embeddings are initialized from other classification tasks. The results in a more fine-grained setting (NP$\rightarrow$P) also indicate the ability of MAML to learn classification tasks. While Reptile performs the best (20.44$\%$) in this setting, MAML still outperforms PT by a large margin (\textbf{11.14}$\%$). 

However, as shown in \Cref{tab:main-result}, MAML falls behind FoMAML when adapting to non-classification tasks. Among the three meta learning methods, FoMAML achieves the best performance (\textbf{9.81}$\%$) on non-classification target tasks in the Both$\rightarrow$Non-Cls setting, showing effective knowledge transfer. We observe similar results in more fine-grained settings QA/Non-QA$\rightarrow$QA, where FoMAML outperforms MAML and Reptile significantly. {While Reptile is claimed empirically to be better than MAML/FoMAML \citep{lee2022meta}, it falls short of MAML/FoMAML in many cases. This might be because MAML and FoMAML are more similar compared to Reptile from a gradient perspective \citep{nichol2018first}.  And since the hyperparameter search is done based on MAML (\cref{sec:hyper}), which means Reptile’s method may be suboptimal.
}

In addition, we can see that meta learning helps PT outperform fine-tuning in several settings including Cls$\rightarrow$Cls (MAML, FoMAML), Both$\rightarrow$Cls (FoMAML) and NP$\rightarrow$P (MAML, Reptile), which demonstrates the superiority of MPT.

\paragraph{$\bullet$ MPT does not always outperform multi-task learning (MTL).} While meta learning is specifically designed for quickly adapting to unseen target tasks, it does not always outperform MTL in PT. From \Cref{tab:main-result}, we can observe that MTL achieves better performance than MPT in many cases, especially on non-classification target tasks. We analyze the reasons as follows:

\begin{itemize}[leftmargin=*,topsep=1pt,itemsep=1pt,parsep=0pt]
    \item Meta learning methods have been shown to be highly sensitive to the hyperparameters \citep{antoniou2018how}, which we could not tune exhaustively due to memory/time constraints {(see Appendix~\ref{sec:hyper_analysis} for hyperparameter sensitivity analysis)}. As mentioned in \cref{sec:hyper}, we select the hyperparameters of MAML using the R$\rightarrow$R setting, and then use the same hyperparameters for all meta learning methods in all settings, which might limit the performance of MPT.
    \item {There might be less shared structure (or features) among non-classification tasks compared to classification. The classification tasks mostly involve sentence-level classification and in some cases the task labels correlate well (\eg\ AG News and DBpedia). Thus, they share some common semantics in both source and target tasks. The model can learn similar patterns (inferring the label of the entire input sentence) during both meta-training and meta-testing stages, enabling better knowledge transfer. The non-classification set on the other hand can include different types of tasks such as QA and summarization; modeling them typically requires a Seq2Seq formulation. These tasks typically lack shared task semantics. For example, the structure of QA is context + question + answer, requiring reasoning ability. In contrast, the structure of summarization is long document + short summary, requiring summarizing ability. Although it has been shown that QA can help summarization in content selection \citep{arumae-liu-2019-guiding}, it is more difficult for MPT to capture transferable knowledge as success of meta learning eventually depends on how much the tasks share \citep{Chelsea2022}.}     
\end{itemize}

{To provide an in-depth analysis of the difference between classification and non-classification tasks, we consider from the perspective of task similarity. We follow \citet{lin2022trgp} which shows that the correlation between input subspaces (the norm of projected subspace onto the other subspace) for two tasks can serve as the similarity score between them. We randomly pick 5 (cls,cls) task pairs as similar tasks. For dissimilar tasks, we randomly pick 5 (QA, summarization) task pairs. The average similarity score for similar task pairs is 0.768 while for dissimilar task pairs the score is only 0.306 {(see Appendix~\ref{sec:task_similarity_analysis} for detailed results)}, which verifies that classification tasks share more structure than non-classification tasks.
}

Given the performance gap between MPT and MTL in some settings, we believe that exploring more advanced MPT methods could be a  promising research direction.

\begin{figure}[t]
    \centering
    \includegraphics[width=0.48\textwidth]{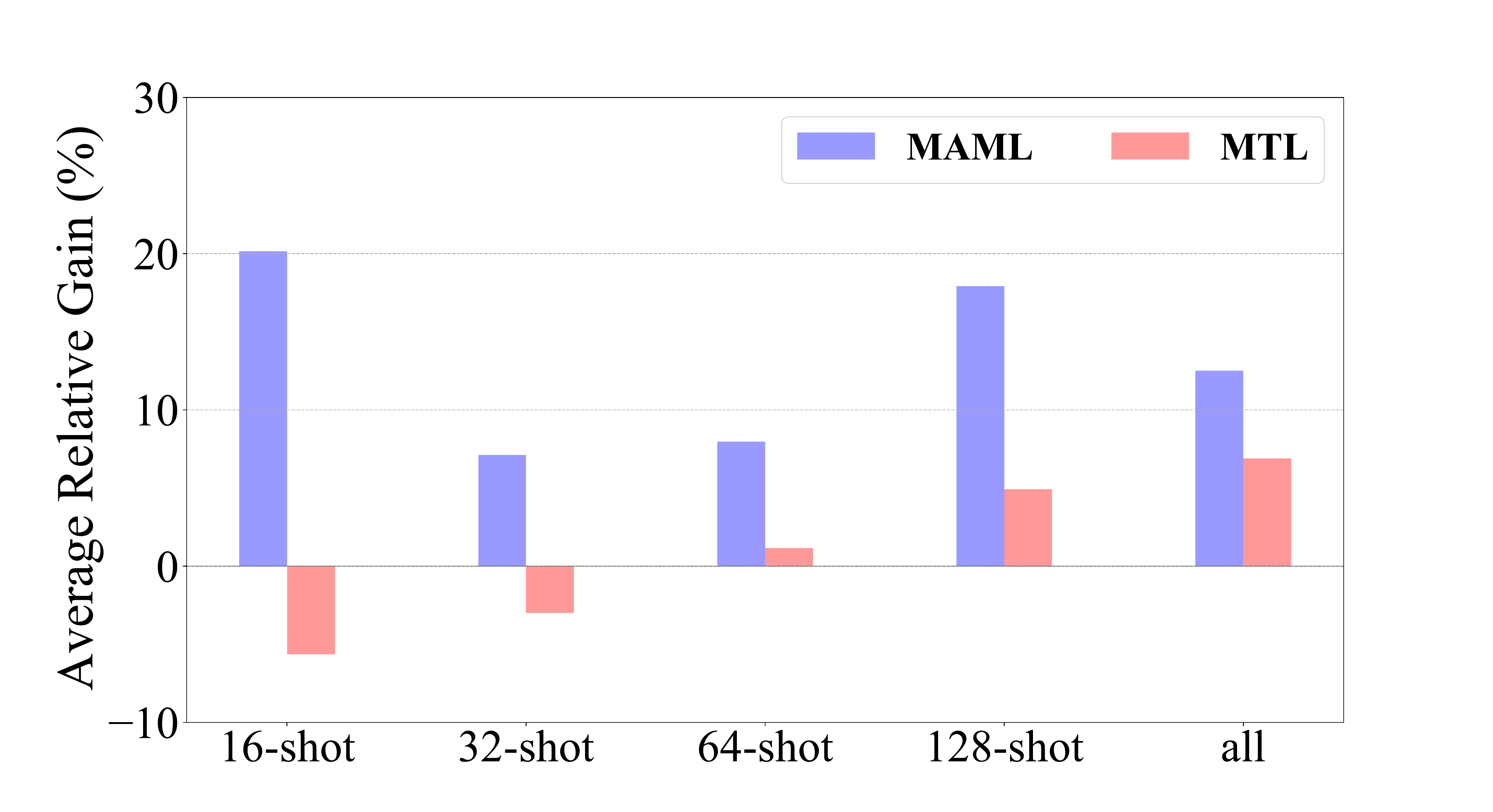}
    \caption{ARG ($\%$) of MPT (MAML) and multi-task learning \wrt\ prompt tuning (ARG = 0) for \textbf{varying data size of source tasks} in the Cls$\rightarrow$Cls setting.}
    \label{fig:more_src}
\end{figure}

\finding{\small \textbf{Q2.} What happens with more labelled data for source/target tasks (beyond few-shot settings)?
}

\noindent As mentioned in \cref{sec:partition}, we mainly explore how MPT improves cross-task generalization when both the source and target tasks are few-shot, which corresponds to the way humans learn \citep{lake_ullman_tenenbaum_gershman_2017}. We used 16 samples per class for classification tasks, and 32 samples per dataset for non-classification tasks. To validate whether more labelled data for source/target tasks can influence the performance of MPT, we conduct controlled experiments with $\{32,64,128,$ all$\}$ samples per class for source/target tasks in the Cls$\rightarrow$Cls setting.

\paragraph{$\bullet$ Source} We report the results of MAML and MTL with more labelled data for the source tasks in \Cref{fig:more_src}. We can observe that: \Ni MPT outperforms PT (ARG = 0) and MTL in all cases including using the full dataset, showing its robustness to data sizes. \Nii Increasing the number of samples in source tasks \emph{does not} necessarily lead to better cross-task generalization for MPT. The best ARG is achieved for 16-shot rather than the full dataset, which justifies using few-shot source tasks. \Niii The performance of MTL improves with more data for source tasks, showing a different learning pattern from MPT. 

\paragraph{$\bullet$ Target} \Cref{tab:more_target} shows the results for increasing the number of examples in target tasks. We can see that: \Ni The performance gain of MPT is evident even using the full dataset ($3.27\%$), demonstrating that it \emph{does} help cross-task generalization beyond few-shot. \Nii MPT outperforms MTL by a large margin in all settings. \Niii MTL is  unstable in terms of ARG scores; while it outperforms PT in 64-shot (1.96$\%$) and all samples ($0.53\%$), it falls behind PT in all other settings, indicating that MPT is a better choice when adapting to classification tasks.

\begin{table}[t]
\centering
    \resizebox{0.82\linewidth}{!}{
    \begin{tabular}{lccccc}
    \toprule
    \multirow{2}{*}{\textbf{Method}} & \multicolumn{5}{c}{\textbf{Shot}} \\
    \cmidrule(lr){2-6}
    & 16 & 32 & 64 & 128 & all \\
    \midrule
    MPT (MAML) & $\mathbf{20.16}$ & $\mathbf{9.10}$ & $\mathbf{5.64}$ & $\mathbf{8.36}$ & $\mathbf{3.27}$ \\
    \midrule
    Multi-task learning & $-5.64$ & $-14.17$ & $1.96$ & $-0.20$ & $0.53$ \\
    
    \bottomrule
    \end{tabular}
    }
\caption{
\label{tab:more_target}
ARG ($\%$) of different methods when \textbf{more labelled data is used in target tasks}.} 
\end{table}

\begin{table}[t]
\centering
    \resizebox{0.72\linewidth}{!}{
    \begin{tabular}{lccc}
    \toprule
    \multirow{2}{*}{\textbf{Method}} & \multicolumn{3}{c}{\textbf{Source task number}} \\
    \cmidrule(lr){2-4}
    & 12 & 24 & 45 \\
    \midrule
    MPT (MAML) & $8.44$ & $12.89$ & $\mathbf{20.16}$   \\
    \bottomrule
    \end{tabular}
    }
\caption{
\label{tab:diff_numbers}
ARG ($\%$) of MPT (MAML) when using different number of source tasks in the Cls$\rightarrow$Cls setting.}
\end{table}

\finding{\small \textbf{Q3.} Does MPT help with more diverse source tasks? \label{Q3}
}
\noindent MPT aims to learn to initialize the prompt embeddings from source tasks, which may cover different types. We hypothesize that the diversity of source tasks might influence its performance. To verify this, we analyze the influence of different source task selections on the same target tasks in two settings: varying the type and number of  tasks.

\paragraph{$\bullet$ Type of tasks.} The results of learning from different types of source tasks are reported in \Cref{tab:main-result}. The performance of MPT on non-classification target tasks improves when using more diverse source tasks, \eg\ from Non-Cls/Cls$\rightarrow$Non-Cls to Both$\rightarrow$Non-Cls. However, for adapting to classification task, the best ARG is achieved when all source tasks are classification, \ie\ the Cls$\rightarrow$Cls setting. Hence, we can conclude that increasing the type diversity of source tasks \emph{does not} necessarily improve cross-task generalization, which is consistent with the finding in \citet{ye-etal-2021-crossfit}.

\paragraph{$\bullet$ Number of tasks.} To investigate the impact of the number of source tasks, we conduct controlled experiments on $\{12,24\}$ source tasks sampled from the original 45 source tasks in the Cls$\rightarrow$Cls setting (see Appendix~\ref{sec:sampled} for a full list). From \Cref{tab:diff_numbers}, we can observe that the performance of MPT keeps improving as the number of source tasks increases, showing better cross-task generalization.

It is worthwhile to note that while our work provides some insights on the choice of source tasks, more systematic studies on how to select the most suitable source tasks given a set of target tasks are needed. We hope that future  analysis can provide a more comprehensive understanding of the relationship between source and target tasks.

\begin{table}[t]
    \centering
    \resizebox{0.95\linewidth}{!}{\begin{tabular}{llccccc}
    \toprule
    \multicolumn{2}{l}{\textbf{Method}} & MAML & FoMAML & Reptile & MTL & Fine-tuning \\
    \cmidrule{3-7}
    \multicolumn{2}{l}{T5-Large} &  $11.14$ & $-0.68$ & $\mathbf{20.44}$ & $-5.19$ & $1.56$ \\
    \multicolumn{2}{l}{T5-Base} &  $\mathbf{9.24}$ & $4.15$ & $7.96$ & $1.64$ & $7.41$ \\
    \multicolumn{2}{l}{T5-XLarge} &  $\mathbf{14.35}$ & $2.46$ & $10.74$ & $5.72$ & $-9.61$ \\
    \multicolumn{2}{l}{BART-Large} &  $7.63$ & $1.16$ & $\mathbf{8.94}$ & $-2.37$ & $2.74$ \\
    \multicolumn{2}{l}{GPT2-Large} &  $3.19$ & $-2.68$ & $\mathbf{4.62}$ & $-1.43$ & $3.75$ \\
    \bottomrule
    \end{tabular}}
    \caption{Average relative gain (ARG $\%$) of all methods with different backbone models in the NP$\rightarrow$P setting. `MTL' stands for `multi-task learning'. 
    }
    \label{tab:t5base}
\end{table}

\finding{\small \textbf{Q4.} Is the performance gain of MPT consistent across different backbone language models?
}
\noindent Our experiments and analysis so far use T5-Large as the backbone model. To verify whether the performance gain of MPT is consistent across different backbone models, we extend the experiments to {T5-Base, T5-XLarge, BART-Large and GPT2-Large} in the NP$\rightarrow$P setting. From the results shown in \Cref{tab:t5base}, we can see that {MPT still outperforms PT and MTL by a large margin when using other PLMs as the backbone model, showing its robustness to model size and type. In addition, the consistent gain of MPT with T5-XLarge could also verify the effectiveness of MPT for huge PLMs which have been shown to perform better in prompt tuning \citep{lester-etal-2021-power}.}

\subsection{Further Analysis} \label{sec:analysis}

\begin{table}[t]
\centering
\resizebox{0.95\linewidth}{!}{
\begin{tabular}{cccc}
\toprule
\textbf{Target Task} & \textbf{Partition} & $\Delta_{\text{MPT}}$ & $\Delta_{\text{MTL}}$  \\
\midrule
\multirow{2}{*}{Amazon\_Polarity}  & R$\rightarrow$R & $3.10$ & $2.25$     \\
& Cls$\rightarrow$Cls & $7.40$ & $10.45$ \\
\midrule
\multirow{2}{*}{AI2\_ARC}  & R$\rightarrow$R & $12.54$ & $5.55$            \\
& Both$\rightarrow$Non-Cls & $8.17$ & $6.69$ \\
\midrule
\multirow{2}{*}{Samsum}  & R$\rightarrow$R & $1.97$ & $6.77$      \\
& Both$\rightarrow$Non-Cls & $2.50$ & $5.71$ \\
\midrule
\multirow{2}{*}{Superglue-Copa}  & Both$\rightarrow$Non-Cls & $1.20$  & $10.00$     \\
& QA$\rightarrow$QA & $-3.20$ & $4.80$ \\
\bottomrule
\end{tabular}
}
\caption{Relative gain in $\%$ for MPT and MTL  when the same target task appears in different patitions.}\label{tab:case}
\end{table}

\paragraph{Prompt tuning (PT) vs. Fine-tuning (FT).} While PT shows strong few-shot learning ability, FT remains the dominant paradigm. As shown in \Cref{tab:main-result}, FT outperforms PT when adapting to classification tasks even in few-shot settings, which might be because PT has only a few tunable parameters. Though MPT is based on PT, its performance gain over FT in all cases suggests that it can learn to initialize the prompt embeddings from source tasks, enabling effective knowledge transfer. 

\paragraph{Case Study} To take a closer look at the influence of different source task types on a particular target task, we further conduct a case study where we ensure that the task under consideration appears in the target task partitions.\footnote{As before, we ensure it does not appear in the source.} Results are shown in \Cref{tab:case}; for example, the first block indicates that \texttt{Amazon\_Polarity} appears as a target task in both R$\rightarrow$R and Cls$\rightarrow$Cls settings. We can observe that there is no consistent conclusion on how we should choose the source tasks for a specific target task, which is consistent with our view in Q3.

\section{Conclusion} \label{sec:conclusion}

In this paper, we have introduced meta prompt tuning (MPT), which learns to initialize the prompt embeddings for adapting to a target task. We have  identified key research questions and systematically studied where and how meta learning can improve cross-task generalization in prompt tuning. We have empirically analyzed a representative set of meta learning methods in a variety of adaptation settings on a large, diverse collection of few-shot tasks. Extensive experimental results and analysis verify the effectiveness of MPT. Given the findings, in the future, we would like to explore more advanced meta learning algorithms which can consistently outperform multi-task learning.

\section*{Limitations}

Although comprehensive, our study of MPT in this work has couple of limitations: 

\begin{itemize}[leftmargin=*,topsep=2pt,itemsep=2pt,parsep=0pt]
    \item As mentioned in \Cref{sec:hyper}, because of infeasiblity to search for optimal hyperparameters for each of the meta learning  methods in each of the ten settings, we choose to use the R$\rightarrow$R setting as our main representative setting. This could be one of the reasons for MPT underperforming MTL in some non-classification tasks (noted in \Cref{sec:exp_res}-Q1).
    \item We mainly focus on how upstream meta learning can improve the performance on target tasks. However, meta learning also enables faster convergence. We leave how it could help reduce the convergence time of PT as future work.
\end{itemize}

Aside from that, meta prompt tuning (MPT) as a method has a limitation that it is {Memory-intensive.}  Optimization-based meta learning methods, especially MAML, are  memory-intensive, which limits the tuning of the inner batch size and inner update steps (\Cref{sec:hyper}). One potential solution is to build more memory-efficient meta learning libraries. 

\bibliography{anthology,custom}
\bibliographystyle{acl_natbib}

\appendix
\section{Appendix}
\label{sec:appendix}

\subsection{Task List} \label{sec:full-data-list}

We report the full list of tasks used in ten different settings in \Cref{tab:full-tasks}. All tasks are taken from CROSSFIT \citep{ye-etal-2021-crossfit}.

\subsection{Relative gain of Every Target Task} \label{sec:detail}
\begin{figure*}[t]
    \centering
    \includegraphics[width=1.00\textwidth]{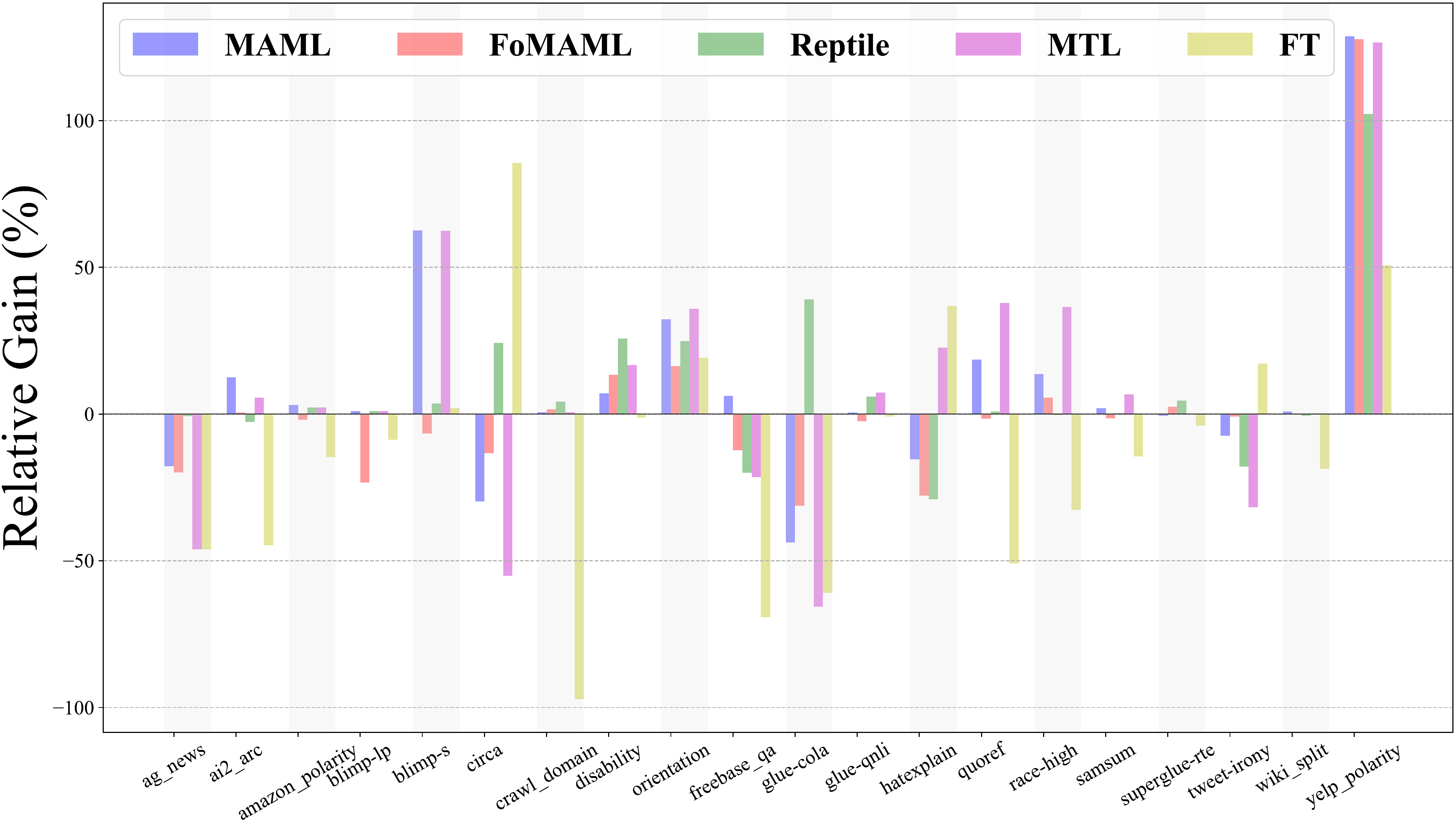}
    \caption{Random to Random (Relative Gain)}
    \label{fig:r2r}
\end{figure*}

\begin{figure*}[t]
    \centering
    \includegraphics[width=1.00\textwidth]{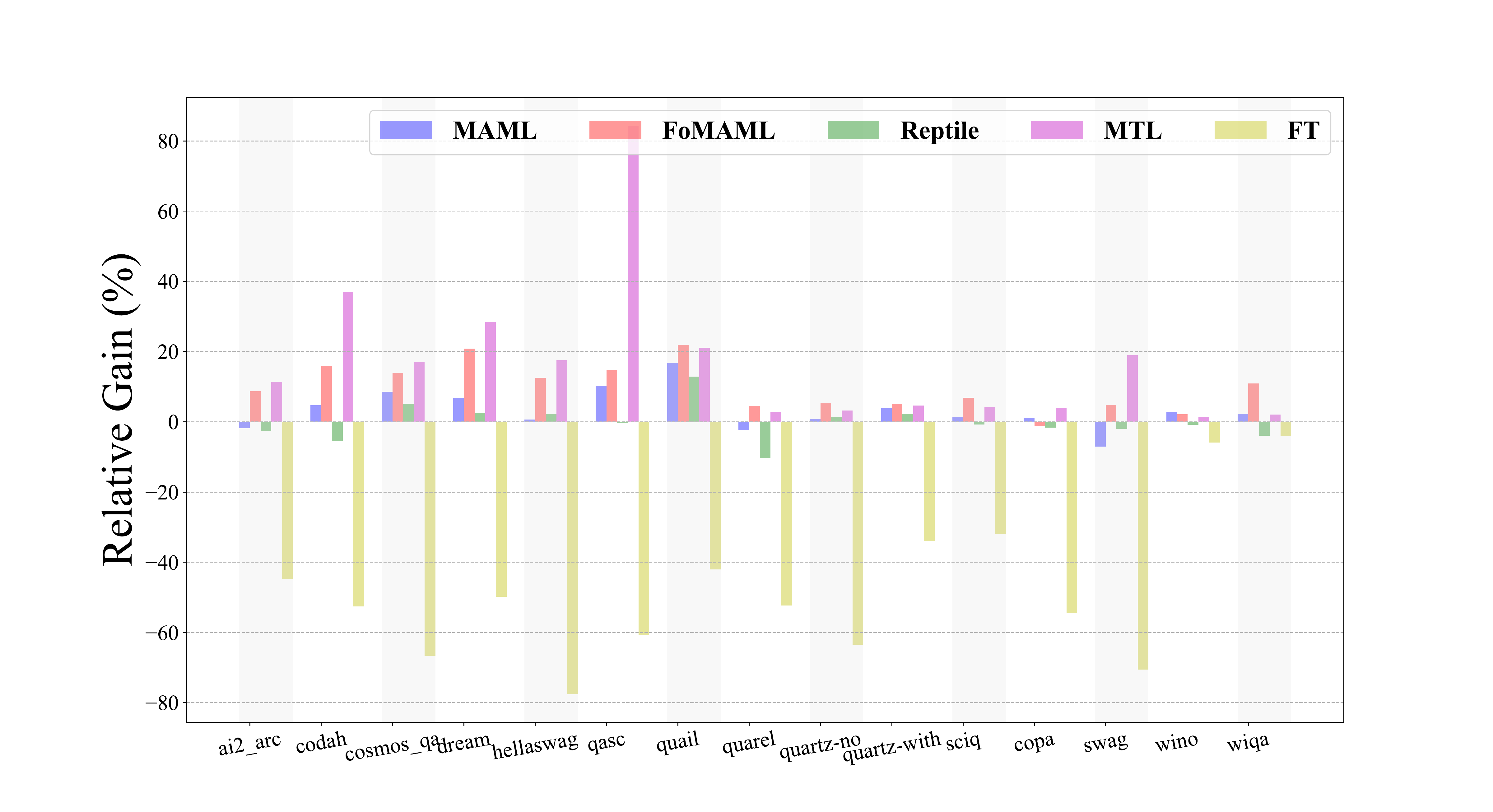}
    \caption{Non-QA to QA (Relative Gain)}
    \label{fig:noqa2qa}
\end{figure*}

We mainly report average relative gain (ARG) in our experiments (\cref{sec:exp_res}). In this section, we show detailed relative gain of each target task in \Cref{fig:r2r} $\sim$ \Cref{fig:noqa2qa}.

\subsection{Absolute Scores for Every Target Task} \label{sec:detail_absolute}
\begin{figure*}[t]
    \centering
    \includegraphics[width=1.00\textwidth]{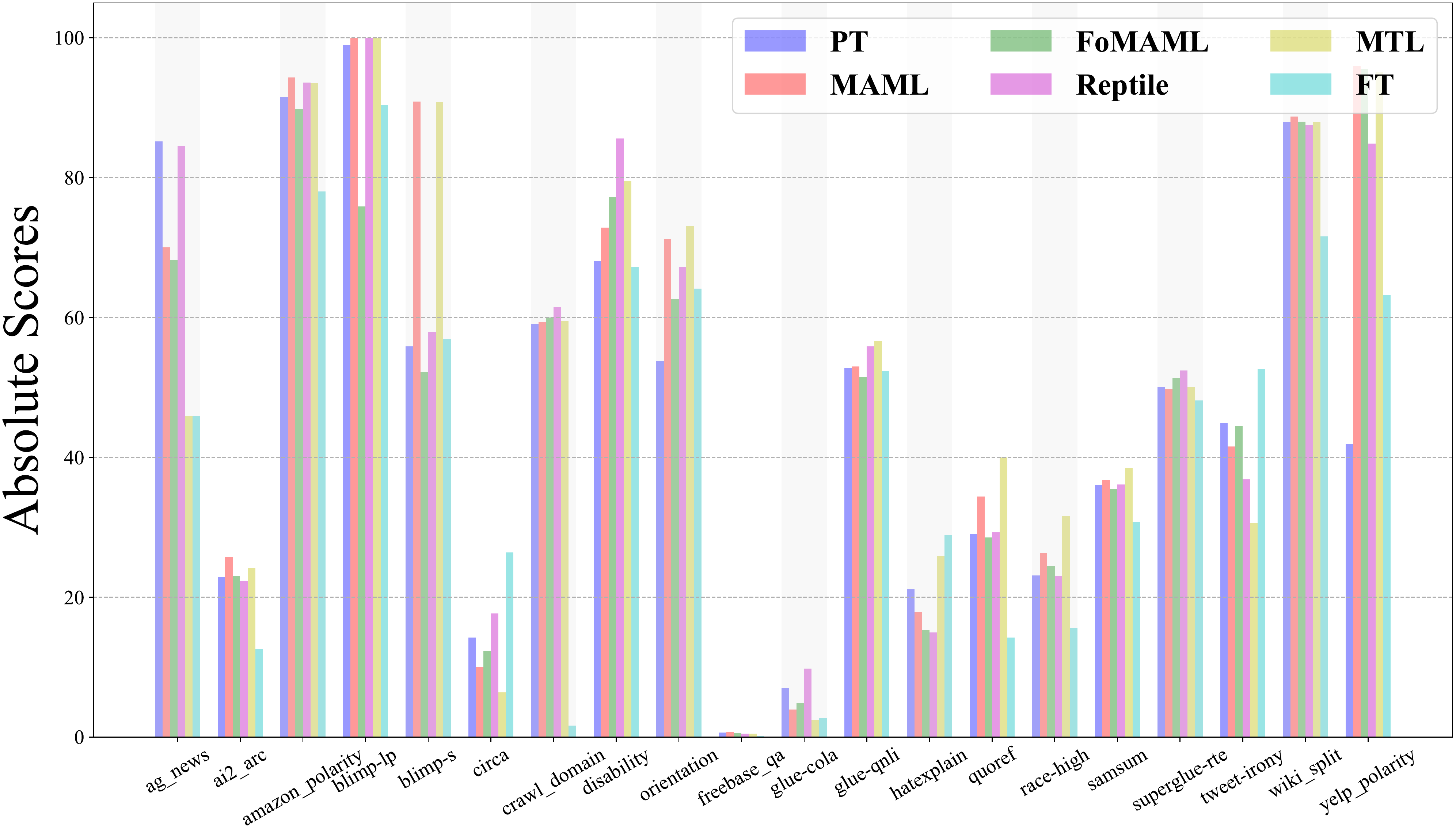}
    \caption{Random to Random (Absolute Scores)}
    \label{fig:r2r_absolute}
\end{figure*}

\begin{figure*}[t]
    \centering
    \includegraphics[width=1.00\textwidth]{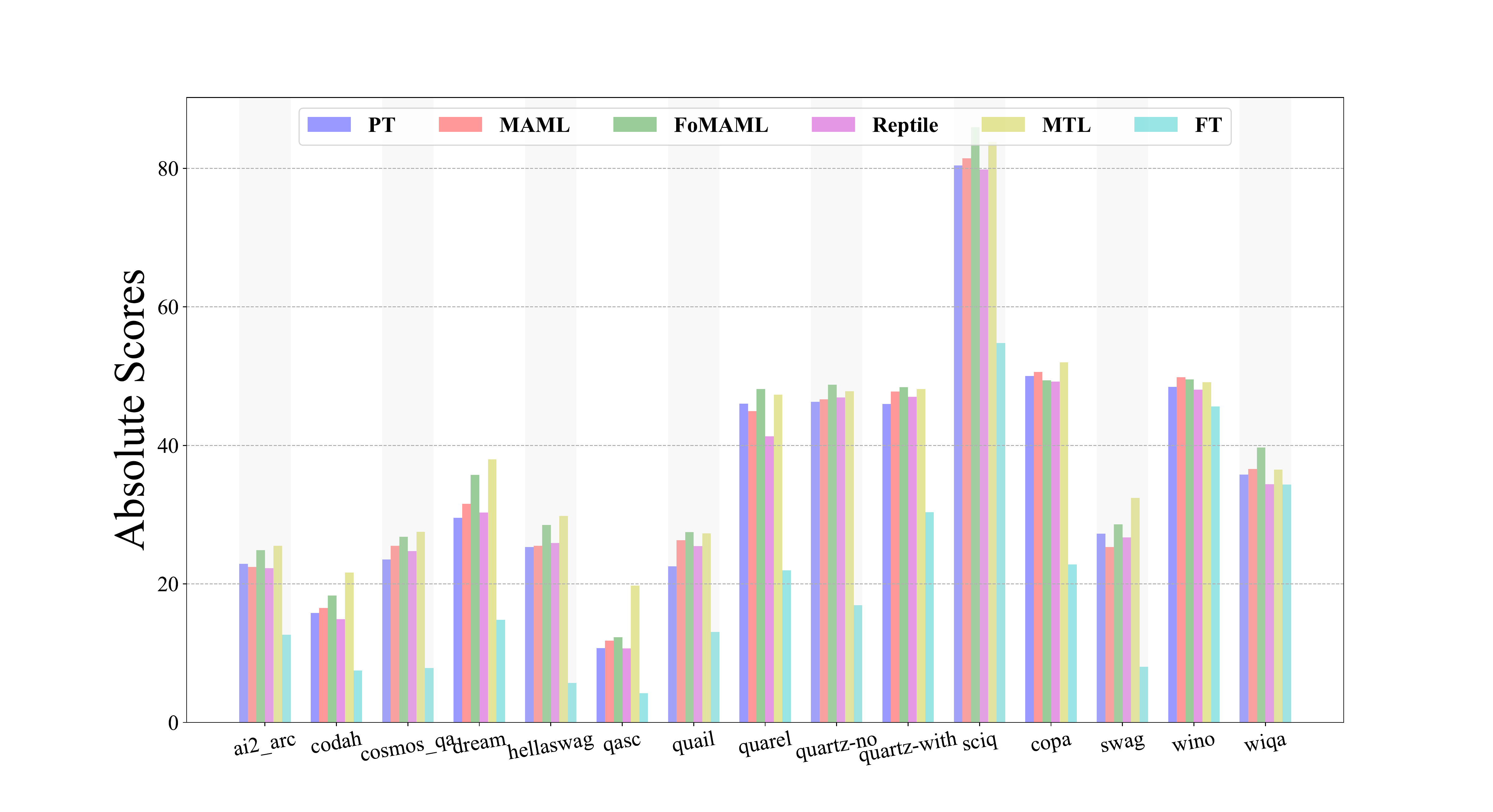}
    \caption{Non-QA to QA (Absolute Scores)}
    \label{fig:noqa2qa_absolute}
\end{figure*}

We show detailed absolute scores for each target task in \Cref{fig:r2r_absolute} $\sim$ \Cref{fig:noqa2qa_absolute}.

\subsection{Details of Sampled Tasks} \label{sec:sampled}
We sample $\{12,24\}$ tasks from the original 45 source tasks in the Cls$\rightarrow$Cls setting to investigate the influence of the number of source tasks. The details of sampled tasks are shown in \Cref{tab:sampled-task}.

\subsection{Hyperparameter Sensitivity Analysis} \label{sec:hyper_analysis}
As mentioned in \Cref{sec:hyper}, for MAML, we select the inner learning rate from $\{2\mathrm{e}{-5},3\mathrm{e}{-5},5\mathrm{e}{-5}\}$, the outer learning rate from $\{2\mathrm{e}{-1},3\mathrm{e}{-1},5\mathrm{e}{-1}\}$, and total training steps from $\{2500,5000,10000\}$ in the R$\rightarrow$R setting. The best validation performance ($10.14\%$ ARG) is achieved with $\{3\mathrm{e}{-5}, 5\mathrm{e}{-1}, 5000\}$, while the worst validation ARG is $-16.21\%$ when using $\{5\mathrm{e}{-5}, 2\mathrm{e}{-1}, 2500\}$. We can see that MPT is quite sensitive to hyperparameters. It performs even worse than PT with inappropriate hyperparameters.

\subsection{Task Similarity Analysis} \label{sec:task_similarity_analysis}
As discussed in \Cref{sec:exp_res}, we use the correlation between input subspaces for two tasks as the similarity score between them. Detailed results of randomly picked similar and dissimilar task pairs are shown in \Cref{tab:all_sim_scores}.

\subsection{Pilot Experiments on Prompt Transfer} \label{sec:pilot_exp}
We conduct some pilot experiments to explore the soft prompt transferability between different source tasks and a given single target task. We randomly pick 3 target tasks in the R$\rightarrow$R setting and conduct prompt tuning on these tasks to obtain their corresponding prompt embeddings $\{P_t^1, P_t^2, P_t^3\}$. We then conduct prompt tuning on 30 randomly selected source tasks to obtain the soft prompts $\{P_s^1,...,P_s^{30}\}$.

As shown in \citet{lin2022trgp}, the correlation between input subspaces (the norm of projected subspace onto the other subspace) for two tasks could serve as the similarity score between them, which may also indicate the transferability. For each source/target task, we regard the soft prompt as the task embedding \citep{zhou2022efficiently} and obtain its subspace by Singular Value Decomposition (SVD) following \citet{saha2021gradient}. We then calculate the correlation scores between a given target task and all source tasks following \citet{lin2022trgp}.

Finally, for each target task, we apply MPT with 3 different sets of source tasks: \Ni 5 source tasks with the highest correlation scores, \Nii 5 randomly picked source tasks, and \Niii 5 source tasks with the lowest correlation scores. The relative gain of every target task is shown in \Cref{tab:prompt_transfer}. We can observe that using 5 source tasks with the highest correlation scores achieves better performance than the other two settings, indicating that input subspaces could be used to measure the soft prompt transferability between different source tasks and a given single target task.

Note that current experiments and analysis are for a single target task. For the average performance of many target tasks, we need more exploration.

\begin{table}[t]
\centering
    \resizebox{1.00\linewidth}{!}{
    \begin{tabular}{lccccccc}
    \toprule
    \multirow{2}{*}{} & \multicolumn{5}{c}{Task Pair Index} & \multirow{2}{*}{Average}\\
    \cmidrule(lr){2-6}
    & 1 & 2 & 3 & 4 & 5 \\
    \midrule
    Similar & 0.772 & 0.695 & 0.754 & 0.819 & 0.802 & 0.768 \\
    \midrule
    Dissimilar & 0.326 & 0.311 & 0.283 & 0.315 & 0.297 & 0.306  \\
    
    \bottomrule
    \end{tabular}
    }
\caption{
\label{tab:all_sim_scores}
Similarity scores of randomly picked similar and dissimilar task pairs.} 
\end{table}

\begin{table}[t]
\centering
    \resizebox{0.90\linewidth}{!}{
    \begin{tabular}{lccc}
    \toprule
    \multirow{2}{*}{\textbf{Target}} & \multicolumn{3}{c}{\textbf{Source}} \\
    \cmidrule(lr){2-4}
    & highest  & random & lowest \\
    \midrule
    Quoref & 7.28 & 3.61 & 0.95   \\
    \midrule
    Glue-Qnli & 9.53 & 4.36 & 4.87  \\
    \midrule
    Samsum & 5.94 & 4.07 & -1.42  \\
    \bottomrule
    \end{tabular}
    }
\caption{
\label{tab:prompt_transfer}
Relative gain in $\%$ for MPT when using different sets of source tasks.} 
\end{table}

\begin{table*}[!t]
    \centering \scriptsize
    \resizebox{1.00\linewidth}{!}{
    \begin{tabular}{p{\textwidth}}
        \toprule
            Partition: \textbf{Random Source} \\
            glue-mrpc, math\_qa, quarel, e2e\_nlg\_cleaned, tweet\_eval-stance\_atheism, lama-squad, tab\_fact, aqua\_rat, tweet\_eval-emoji, glue-wnli,
            codah, tweet\_eval-offensive, wiki\_qa, blimp-ellipsis\_n\_bar\_1, openbookqa, sms\_spam, acronym\_identification, blimp-determiner\_noun\_agreement\_with\_adj\_irregular\_1, ethos-national\_origin, spider,
            hellaswag, superglue-wsc, numer\_sense, ade\_corpus\_v2-dosage, blimp-ellipsis\_n\_bar\_2, kilt\_ay2, squad-no\_context, google\_wellformed\_query, xsum, wiqa,
            tweet\_eval-stance\_abortion, reddit\_tifu-tldr, ade\_corpus\_v2-effect, qa\_srl, ethos-religion, commonsense\_qa, biomrc, superglue-multirc, ethos-race, eli5-askh,
            glue-qqp, paws, ethos-directed\_vs\_generalized, glue-sst2, tweet\_eval-hate, glue-rte, blimp-anaphor\_number\_agreement, lama-conceptnet, hate\_speech\_offensive, superglue-wic,
            boolq, kilt\_hotpotqa, quartz-no\_knowledge, aslg\_pc12, sick, tweet\_eval-stance\_climate, tweet\_eval-sentiment, crows\_pairs, glue-mnli, medical\_questions\_pairs,
            break-QDMR-high-level, qasc, imdb, ethos-gender, trec-finegrained, adversarialqa, onestop\_english, web\_questions, duorc, swag,
            proto\_qa, scitail, tweet\_eval-stance\_feminist, limit, common\_gen, scicite, blimp-irregular\_past\_participle\_adjectives, social\_i\_qa, anli, kilt\_zsre,
            cosmos\_qa, superglue-record, squad-with\_context, emotion, blimp-existential\_there\_quantifiers\_1, race-middle, kilt\_wow, sciq, wino\_grande, rotten\_tomatoes,
            superglue-cb, poem\_sentiment, ropes, reddit\_tifu-title, piqa, climate\_fever, lama-google\_re, search\_qa, mc\_taco, blimp-wh\_questions\_object\_gap,
            hotpot\_qa, emo, kilt\_nq, kilt\_trex, quartz-with\_knowledge, dbpedia\_14, yahoo\_answers\_topics, superglue-copa, blimp-anaphor\_gender\_agreement, hate\_speech18,
            gigaword, multi\_news, aeslc, quail \\
        \midrule
            Partition: \textbf{Random Target} \\
            quoref, wiki\_split, ethos-disability, yelp\_polarity, superglue-rte,
            glue-cola, ethos-sexual\_orientation, blimp-sentential\_negation\_npi\_scope,
            ai2\_arc, amazon\_polarity, race-high, blimp-sentential\_negation\_npi\_licensor\_present,
            tweet\_eval-irony, crawl\_domain, freebase\_qa, glue-qnli,
            hatexplain, ag\_news, circa, samsum \\
        \midrule
            Partition: \textbf{Classification Source} \\
            superglue-rte, tweet\_eval-sentiment, discovery, glue-rte, superglue-wsc, scicite, glue-mrpc, tweet\_eval-stance\_hillary, tweet\_eval-offensive, emotion,
            hatexplain, glue-cola, sick, paws, ethos-sexual\_orientation, glue-qqp, tweet\_eval-emotion, sms\_spam, health\_fact, glue-mnli,
            imdb, ethos-disability, glue-wnli, scitail, trec-finegrained, yahoo\_answers\_topics, liar, glue-sst2, tweet\_eval-stance\_abortion, circa,
            tweet\_eval-stance\_climate, glue-qnli, tweet\_eval-emoji, ethos-directed\_vs\_generalized, ade\_corpus\_v2-classification, ag\_news, hate\_speech\_offensive, superglue-wic, google\_wellformed\_query, tweet\_eval-irony,
            ethos-gender, onestop\_english, trec, rotten\_tomatoes, kilt\_fever \\
        \midrule
            Partition: \textbf{Non-Classification Source} \\
            ade\_corpus\_v2-dosage, art, biomrc, blimp-anaphor\_number\_agreement, blimp-ellipsis\_n\_bar\_2, blimp-sentential\_negation\_npi\_licensor\_present, blimp-sentential\_negation\_npi\_scope, break-QDMR-high-level, commonsense\_qa, crows\_pairs,
            dream, duorc, eli5-asks, eli5-eli5, freebase\_qa, gigaword, hellaswag, hotpot\_qa, kilt\_ay2, kilt\_hotpotqa,
            kilt\_trex, kilt\_zsre, lama-conceptnet, lama-google\_re, lama-squad, math\_qa, numer\_sense, openbookqa, piqa, proto\_qa,
            qa\_srl, quarel, quartz-no\_knowledge, race-high, reddit\_tifu-title, reddit\_tifu-tldr, ropes, sciq, social\_i\_qa, spider,
            superglue-multirc, wiki\_bio, wikisql, xsum, yelp\_review\_full
            \\
        \midrule
            Partition: \textbf{Both (Classification + Non-Classification) Source} \\
            ade\_corpus\_v2-dosage, biomrc, blimp-ellipsis\_n\_bar\_2, blimp-sentential\_negation\_npi\_scope, commonsense\_qa,
            crows\_pairs, duorc, hellaswag, kilt\_zsre, lama-google\_re,
            lama-squad, math\_qa, numer\_sense, openbookqa, piqa, proto\_qa, quartz-no\_knowledge, race-high, reddit\_tifu-tldr, ropes,
            sciq, wiki\_bio, discovery, emotion, ethos-disability, ethos-sexual\_orientation, glue-cola, glue-mnli, glue-mrpc, glue-qqp,
            glue-rte, glue-wnli, hatexplain, health\_fact, imdb, paws, scicite, sick, sms\_spam, superglue-rte, superglue-wsc, tweet\_eval-emotion, tweet\_eval-offensive, tweet\_eval-sentiment, tweet\_eval-stance\_hillary
            \\
        \midrule
            Partition: \textbf{Classification Target} \\
            superglue-cb,dbpedia\_14,wiki\_qa,emo,yelp\_polarity,ethos-religion,amazon\_polarity,tab\_fact,anli,ethos-race \\
        \midrule
            Partition: \textbf{Non-Classification Target} \\
            multi\_news, superglue-copa, quail, blimp-anaphor\_gender\_agreement,
            common\_gen, acronym\_identification, quoref, wiki\_split,
            ai2\_arc, break-QDMR, crawl\_domain, samsum \\
        \midrule
            Partition: \textbf{QA Source} \\
            biomrc, boolq, freebase\_qa, hotpot\_qa, kilt\_hotpotqa, kilt\_nq, kilt\_trex, kilt\_zsre, lama-conceptnet, lama-google\_re,
            lama-squad, lama-trex, mc\_taco, numer\_sense, quoref, ropes, search\_qa, squad-no\_context, superglue-multirc,
            superglue-record, tweet\_qa, web\_questions \\
        \midrule
            Partition: \textbf{Non-QA Source} \\
            hate\_speech\_offensive, google\_wellformed\_query, circa, glue-sst2, scitail,
            emo, ag\_news, art, paws, kilt\_ay2,
            glue-qnli, ade\_corpus\_v2-classification, hatexplain, emotion, glue-qqp,
            kilt\_fever, dbpedia\_14, glue-mnli, discovery, gigaword,
            amazon\_polarity, tab\_fact, tweet\_eval-emoji, tweet\_eval-offensive, tweet\_eval-sentiment,
            imdb, liar, anli, wikisql, xsum,
            yahoo\_answers\_topics, yelp\_polarity, yelp\_review\_full\\ 
        \midrule
            Partition: \textbf{QA Target} \\
            ai2\_arc, codah, cosmos\_qa, dream,
            hellaswag, qasc, quail,
            quarel, quartz-no\_knowledge, quartz-with\_knowledge, sciq,
            superglue-copa, swag, wino\_grande, wiqa\\
        \midrule
            Partition: \textbf{Non-Paraphrase Classification Source} \\
            ade\_corpus\_v2-classification, ag\_news, amazon\_polarity, anli, circa, climate\_fever, dbpedia\_14, discovery, emo, emotion, ethos-directed\_vs\_generalized, ethos-disability, ethos-gender, ethos-national\_origin, ethos-race, ethos-religion, ethos-sexual\_orientation, financial\_phrasebank, glue-cola, glue-mnli, glue-qnli, glue-rte, glue-sst2, glue-wnli, google\_wellformed\_query, hate\_speech18, hate\_speech\_offensive, hatexplain, health\_fact, imdb, kilt\_fever, liar, onestop\_english, poem\_sentiment, rotten\_tomatoes, scicite, scitail, sick, sms\_spam, superglue-cb, superglue-rte, superglue-wic, superglue-wsc, tab\_fact, trec, trec-finegrained, tweet\_eval-emoji, tweet\_eval-emotion, tweet\_eval-hate, tweet\_eval-irony, tweet\_eval-offensive, tweet\_eval-sentiment, tweet\_eval-stance\_abortion, tweet\_eval-stance\_atheism, tweet\_eval-stance\_climate, tweet\_eval-stance\_feminist, tweet\_eval-stance\_hillary, wiki\_qa, yahoo\_answers\_topics, yelp\_polarity
            \\
        \midrule
            Partition: \textbf{Paraphrase Target} \\
            glue-mrpc, glue-qqp, medical\_questions\_pairs, paws
            \\
        \bottomrule
    \end{tabular}
    }
    \caption{Full datasets for all settings described in Section~\ref{sec:partition}. We provide references for all datasets in \Cref{tab:full-citations}.
    }\label{tab:full-tasks}
\end{table*}

\begin{table*}[!t]
    \centering \scriptsize
    \resizebox{1.00\linewidth}{!}{
    \begin{tabular}{p{\textwidth}}
        \midrule
            \textbf{12 source tasks} \\
            superglue-rte, tweet\_eval-sentiment, discovery, glue-rte, hatexplain, glue-cola, health\_fact, glue-mnli, imdb, ethos-disability, glue-wnli, scitail \\
        \midrule
            \textbf{24 source tasks} \\
            superglue-rte, tweet\_eval-sentiment, discovery, glue-rte, superglue-wsc, scicite, hatexplain, glue-cola, tweet\_eval-emotion, sms\_spam, health\_fact, glue-mnli, imdb, ethos-disability, glue-wnli, scitail, glue-sst2, tweet\_eval-stance\_abortion, glue-qnli, ethos-directed\_vs\_generalized, ag\_news, hate\_speech\_offensive, ethos-gender, kilt\_fever
            \\
        \bottomrule
    \end{tabular}
    }
    \caption{Details of sampled $\{12,24\}$ tasks for investigating the impact of the number of source tasks.
    }\label{tab:sampled-task}
\end{table*}

\clearpage

\scriptsize
\onecolumn
\begin{longtable}{ll}
    \toprule
    \textbf{Task Name}  & \textbf{Reference} \\
    \midrule
    \endfirsthead
    \toprule
    \textbf{Task Name}  & \textbf{Reference} \\
    \midrule
    \endhead
    \bottomrule\\
    \endfoot
    \endlastfoot
        
        eli5-eli5  & \citealt{fan-etal-2019-eli5}    \\
        ethos-race  &   \citealt{Mollas2020ETHOSAO}     \\
        tweet\_qa  &    \citealt{xiong-etal-2019-tweetqa}       \\
        tweet\_eval-stance\_hillary  &  \citealt{barbieri-etal-2020-tweeteval}  \\
        piqa &  \citealt{Bisk2020}      \\
        acronym\_identification  &      \citealt{pouran-ben-veyseh-etal-2020-acronym}   \\
        wiki\_split  & \citealt{botha-etal-2018-learning}       \\
        scitail  & \citealt{scitail}    \\
        emotion  & \citealt{saravia-etal-2018-carer}    \\
        medical\_questions\_pairs  & \citealt{medical-qqp}      \\
        blimp-anaphor\_gender\_agreement  & \citealt{warstadt2019blimp} \\
        sciq  & \citealt{welbl-etal-2017-crowdsourcing} \\
        paws  & \citealt{zhang-etal-2019-paws}  \\
        yelp\_review\_full  &   \citealt{zhang2015character}; \href{https://www.yelp.com/dataset}{(link)}       \\
        freebase\_qa  & \citealt{jiang-etal-2019-freebaseqa}    \\
        anli  & \citealt{nie-etal-2020-adversarial}     \\
        quartz-with\_knowledge  &       \citealt{tafjord-etal-2019-quartz}      \\
        hatexplain  &   \citealt{mathew2020hatexplain}  \\
        yahoo\_answers\_topics  &       \href{https://webscope.sandbox.yahoo.com/catalog.php?datatype=l}{(link)}        \\
        search\_qa  &   \citealt{Dunn2017SearchQAAN}    \\
        tweet\_eval-stance\_feminist  & \citealt{barbieri-etal-2020-tweeteval}  \\
        codah  &        \citealt{chen-etal-2019-codah}  \\
        lama-squad  &   \citealt{petroni-etal-2019-language,petroni2020how}     \\
        superglue-record  & \citealt{Zhang2018ReCoRDBT} \\
        spider  &       \citealt{yu-etal-2018-spider}   \\
        mc\_taco  &     \citealt{zhou-etal-2019-going}  \\
        glue-mrpc  & \citealt{dolan-brockett-2005-automatically}\\
        kilt\_fever  &  \citealt{thorne-etal-2018-fever}        \\
        eli5-asks  qa & \citealt{fan-etal-2019-eli5}    \\
        imdb  & \citealt{maas-etal-2011-learning}       \\
        tweet\_eval-stance\_abortion  & \citealt{barbieri-etal-2020-tweeteval}  \\
        aqua\_rat  &    \citealt{ling-etal-2017-program}        \\
        duorc  &        \citealt{saha-etal-2018-duorc}  \\
        lama-trex  &    \citealt{petroni-etal-2019-language,petroni2020how}     \\
        tweet\_eval-stance\_atheism  &  \citealt{barbieri-etal-2020-tweeteval}  \\
        ropes  &        \citealt{lin-etal-2019-reasoning}       \\
        squad-no\_context  &    \citealt{rajpurkar-etal-2016-squad}     \\
        superglue-rte  & \citealt{dagan2005pascal}  \\
        qasc  & \citealt{Khot_Clark_Guerquin_Jansen_Sabharwal_2020}     \\
        hate\_speech\_offensive  &      \citealt{hateoffensive} \\
        trec-finegrained  &     \citealt{li-roth-2002-learning,hovy-etal-2001-toward}   \\
        glue-wnli  & \citealt{levesque2012winograd}     \\
        yelp\_polarity  & \citealt{zhang2015character}; \href{https://www.yelp.com/dataset}{(link)}     \\
        kilt\_hotpotqa  &       \citealt{yang-etal-2018-hotpotqa}       \\
        glue-sst2  &    \citealt{socher-etal-2013-recursive}    \\
        xsum  & \citealt{narayan-etal-2018-dont}        \\
        tweet\_eval-offensive  &        \citealt{barbieri-etal-2020-tweeteval}  \\
        aeslc  & \citealt{zhang-tetreault-2019-email}   \\
        emo  & \citealt{chatterjee-etal-2019-semeval}   \\
        hellaswag  &    \citealt{zellers-etal-2019-hellaswag}   \\
        social\_i\_qa  &        \citealt{sap-etal-2019-social}  \\
        kilt\_wow  &    \citealt{dinan2018wizard}       \\
        scicite  &      \citealt{cohan-etal-2019-structural}    \\
        superglue-wsc  &        \citealt{levesque2012winograd}  \\
        hate\_speech18  &       \citealt{gibert2018hate}        \\
        adversarialqa  &        \citealt{bartolo-etal-2020-beat}        \\
        break-QDMR &    \citealt{wolfson-etal-2020-break}       \\
        dream  &        \citealt{sun-etal-2019-dream}   \\
        circa  &        \citealt{louis-etal-2020-id}    \\
        wiki\_qa  &     \citealt{yang-etal-2015-wikiqa} \\
        ethos-directed\_vs\_generalized  &      \citealt{Mollas2020ETHOSAO}     \\
        wiqa  & \citealt{tandon-etal-2019-wiqa} \\
        poem\_sentiment  & \citealt{sheng-uthus-2020-investigating}     \\
        kilt\_ay2 &     \citealt{hoffart-etal-2011-robust}      \\
        cosmos\_qa  &   \citealt{huang-etal-2019-cosmos}        \\
        reddit\_tifu-title  &   \citealt{kim-etal-2019-abstractive}     \\
        superglue-cb  & \citealt{Marneffe_Simons_Tonhauser_2019}        \\
        kilt\_nq  &     \citealt{kwiatkowski-etal-2019-natural} \\
        quarel  &       \citealt{Tafjord_Clark_Gardner_Yih_Sabharwal_2019}      \\
        race-high  &    \citealt{lai-etal-2017-race}    \\
        wino\_grande  & \citealt{Sakaguchi_Le_Bras_Bhagavatula_Choi_2020}       \\
        break-QDMR-high-level & \citealt{wolfson-etal-2020-break}       \\
        tweet\_eval-irony  &    \citealt{barbieri-etal-2020-tweeteval}  \\
        liar  & \citealt{wang-2017-liar}        \\
        openbookqa  &   \citealt{mihaylov-etal-2018-suit}       \\
        superglue-multirc  &    \citealt{khashabi-etal-2018-looking}    \\
        race-middle  &  \citealt{lai-etal-2017-race}    \\
        quoref  &       \citealt{dasigi-etal-2019-quoref}       \\
        cos\_e  & \citealt{rajani-etal-2019-explain}    \\
        reddit\_tifu-tldr  &    \citealt{kim-etal-2019-abstractive}     \\
        ai2\_arc  &     \citealt{Clark2018ThinkYH}      \\
        quail  &        \citealt{Rogers_Kovaleva_Downey_Rumshisky_2020} \\
        crawl\_domain & \citealt{zhang-etal-2020-semi}  \\
        glue-cola  &    \citealt{warstadt-etal-2019-neural}     \\
        art &   \citealt{bhagavatula2020abductive}      \\
        rotten\_tomatoes  & \citealt{pang-lee-2005-seeing}      \\
        tweet\_eval-emoji  & \citealt{barbieri-etal-2020-tweeteval}     \\
        numer\_sense  & \citealt{lin-etal-2020-birds}   \\
        blimp-existential\_there\_quantifiers\_1  & \citealt{warstadt2019blimp} \\
        eli5-askh  qa & \citealt{fan-etal-2019-eli5}    \\
        ethos-national\_origin  &       \citealt{Mollas2020ETHOSAO}     \\
        boolq  &        \citealt{clark-etal-2019-boolq} \\
        qa\_srl &       \citealt{he-etal-2015-question} \\
        sms\_spam  &    \citealt{sms_spam}      \\
        samsum  &       \citealt{gliwa-etal-2019-samsum}        \\
        ade\_corpus\_v2-classification  &       \citealt{GURULINGAPPA2012885}   \\
        superglue-wic  &    \citealt{pilehvar-camacho-collados-2019-wic}    \\
        ade\_corpus\_v2-dosage  & \citealt{GURULINGAPPA2012885} \\
        tweet\_eval-stance\_climate  &  \citealt{barbieri-etal-2020-tweeteval}  \\
        e2e\_nlg\_cleaned &     \citealt{dusek.etal2020:csl, dusek-etal-2019-semantic}  \\
        aslg\_pc12 &    \citealt{Othman2012EnglishASLGP}        \\
        ag\_news  &     \href{http://groups.di.unipi.it/~gulli/AG_corpus_of_news_articles.html}{Gulli (link)}   \\
        math\_qa  &     \citealt{amini-etal-2019-mathqa}        \\
        commonsense\_qa  &      \citealt{talmor-etal-2019-commonsenseqa}        \\
        web\_questions  &       \citealt{berant-etal-2013-semantic}     \\
        biomrc  &       \citealt{pappas-etal-2020-biomrc}       \\
        swag  & \citealt{zellers-etal-2018-swag}        \\
        blimp-determiner\_noun\_agreement\_with\_adj\_irregular\_1  & \citealt{warstadt2019blimp}       \\
        glue-mnli  & \citealt{williams-etal-2018-broad} \\
        squad-with\_context  &  \citealt{rajpurkar-etal-2016-squad}     \\
        blimp-ellipsis\_n\_bar\_2  & \citealt{warstadt2019blimp}        \\
        financial\_phrasebank  &        \citealt{financial-phrasebank}  \\
        sick  & \citealt{marelli-etal-2014-sick}        \\
        ethos-religion  &       \citealt{Mollas2020ETHOSAO}     \\
        hotpot\_qa  &   \citealt{yang-etal-2018-hotpotqa}       \\
        tweet\_eval-emotion  &  \citealt{barbieri-etal-2020-tweeteval}  \\
        dbpedia\_14  &  \citealt{Lehmann2015DBpediaA}   \\
        ethos-gender  & \citealt{Mollas2020ETHOSAO}     \\
        tweet\_eval-hate  &     \citealt{barbieri-etal-2020-tweeteval}  \\
        ethos-sexual\_orientation  &    \citealt{Mollas2020ETHOSAO}     \\
        health\_fact  & \citealt{kotonya-toni-2020-explainable-automated}       \\
        common\_gen &   \citealt{lin-etal-2020-commongen}       \\
        crows\_pairs &  \citealt{nangia-etal-2020-crows}        \\
        ade\_corpus\_v2-effect  & \citealt{GURULINGAPPA2012885} \\
        blimp-sentential\_negation\_npi\_scope  & \citealt{warstadt2019blimp}   \\
        lama-conceptnet  &      \citealt{petroni-etal-2019-language,petroni2020how}     \\
        glue-qnli  &    \citealt{rajpurkar-etal-2016-squad}     \\
        quartz-no\_knowledge  & \citealt{tafjord-etal-2019-quartz}      \\
        google\_wellformed\_query  &    \citealt{faruqui-das-2018-identifying}  \\
        kilt\_trex  &   \citealt{elsahar-etal-2018-rex} \\
        blimp-ellipsis\_n\_bar\_1  & \citealt{warstadt2019blimp}        \\
        trec  & \citealt{li-roth-2002-learning,hovy-etal-2001-toward}   \\
        superglue-copa  &       \citealt{gordon-etal-2012-semeval} \\
        ethos-disability  &     \citealt{Mollas2020ETHOSAO}     \\
        lama-google\_re  &      \citealt{petroni-etal-2019-language,petroni2020how}     \\
        discovery  &    \citealt{sileo-etal-2019-mining}        \\
        blimp-anaphor\_number\_agreement  & \citealt{warstadt2019blimp} \\
        climate\_fever  &       \citealt{Diggelmann2020CLIMATEFEVERAD}  \\
        blimp-irregular\_past\_participle\_adjectives  & \citealt{warstadt2019blimp}    \\
        tab\_fact  &    \citealt{Chen2020TabFact}       \\
        gigaword  &     \citealt{napoles-etal-2012-annotated}   \\
        glue-rte  &     \citealt{dagan2005pascal} \\
        tweet\_eval-sentiment  &        \citealt{barbieri-etal-2020-tweeteval}  \\
        limit & \citealt{manotas-etal-2020-limit}       \\
        wikisql  &      \citealt{zhongSeq2SQL2017}      \\
        glue-qqp  & \href{http://data.quora.com/First-Quora-Dataset-Release-Question-Pairs}{(link)}     \\
        onestop\_english  &     \citealt{vajjala-lucic-2018-onestopenglish}     \\
        amazon\_polarity  & \citealt{McAuley2013HiddenFA}       \\
        blimp-wh\_questions\_object\_gap  & \citealt{warstadt2019blimp} \\
        multi\_news  & \citealt{fabbri-etal-2019-multi} \\
        proto\_qa &     \citealt{boratko-etal-2020-protoqa}     \\
        wiki\_bio  &    \citealt{lebret-etal-2016-neural}       \\
        kilt\_zsre  &   \citealt{levy-etal-2017-zero}   \\
        blimp-sentential\_negation\_npi\_licensor\_present  & \citealt{warstadt2019blimp}       \\
\bottomrule
\caption{References for all datasets.}\label{tab:full-citations}\\
\end{longtable}
\twocolumn
\normalsize

\begin{figure}[t]
    \centering
    \includegraphics[width=0.48\textwidth]{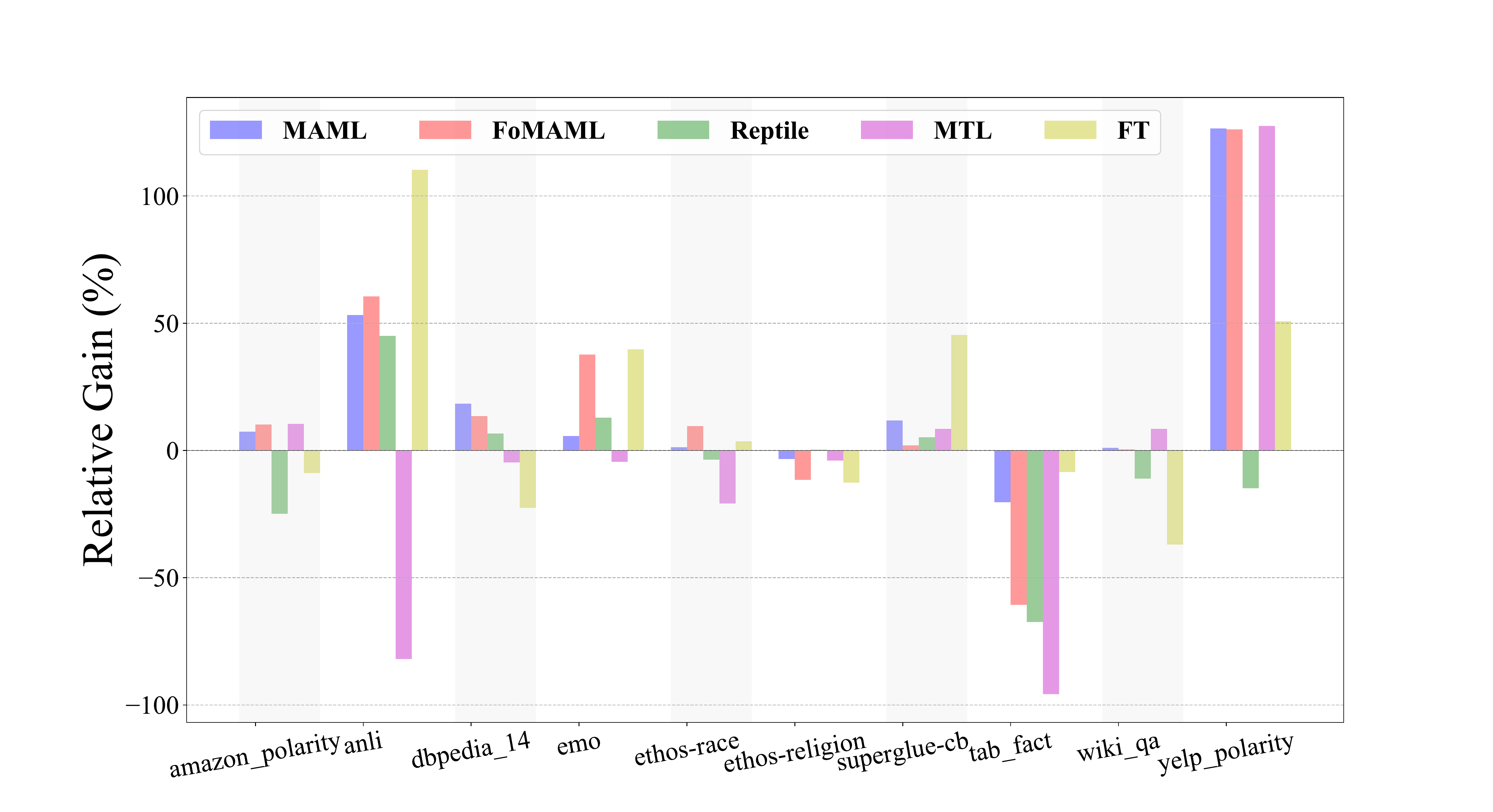}
    \caption{Classification to Classification (Relative Gain)}
    \label{fig:cls2cls}
\end{figure}

\begin{figure}[t]
    \centering
    \includegraphics[width=0.48\textwidth]{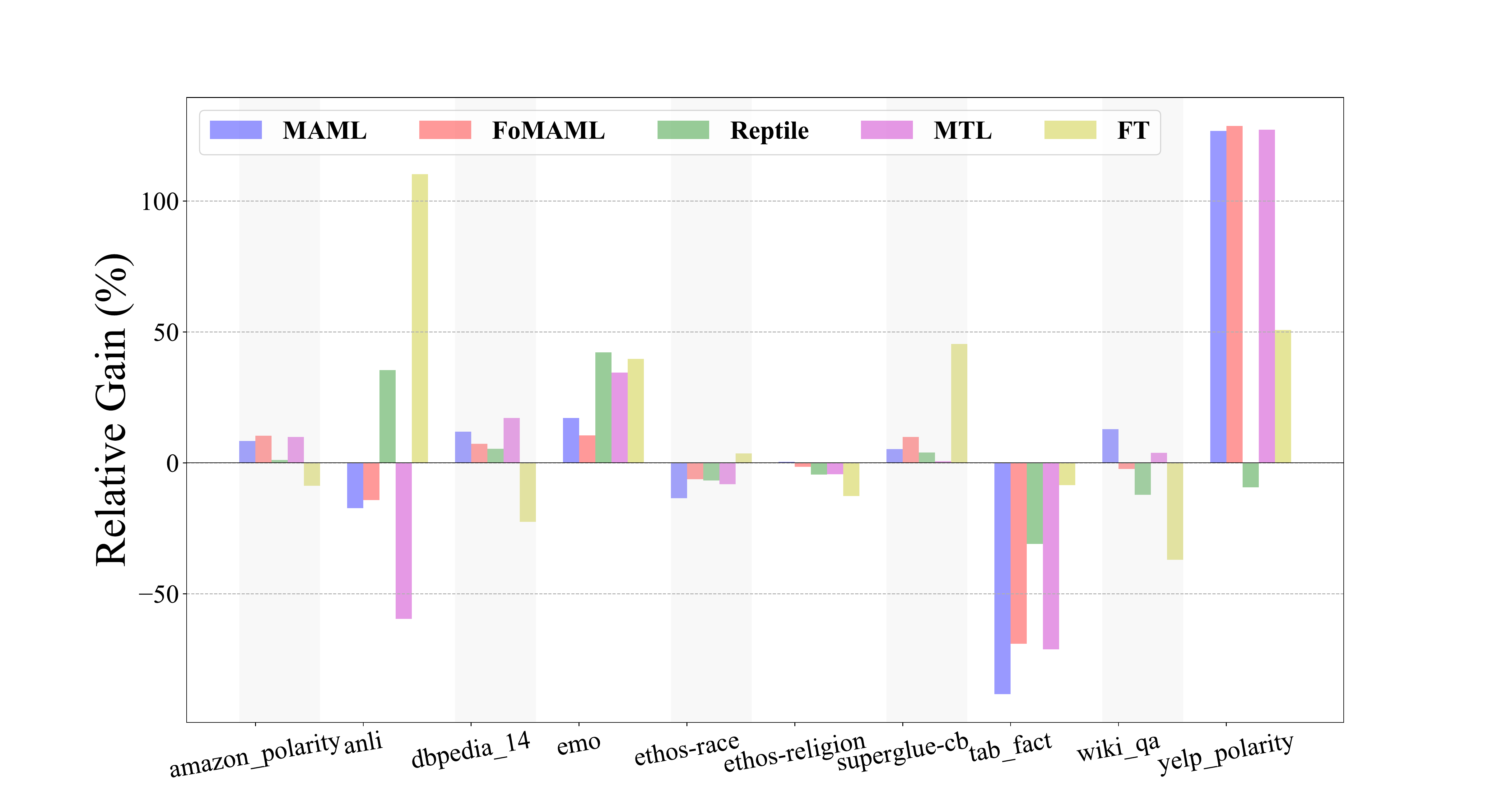}
    \caption{Non-Classification to Classification (Relative Gain)}
    \label{fig:nocls2cls}
\end{figure}

\begin{figure}[t]
    \centering
    \includegraphics[width=0.48\textwidth]{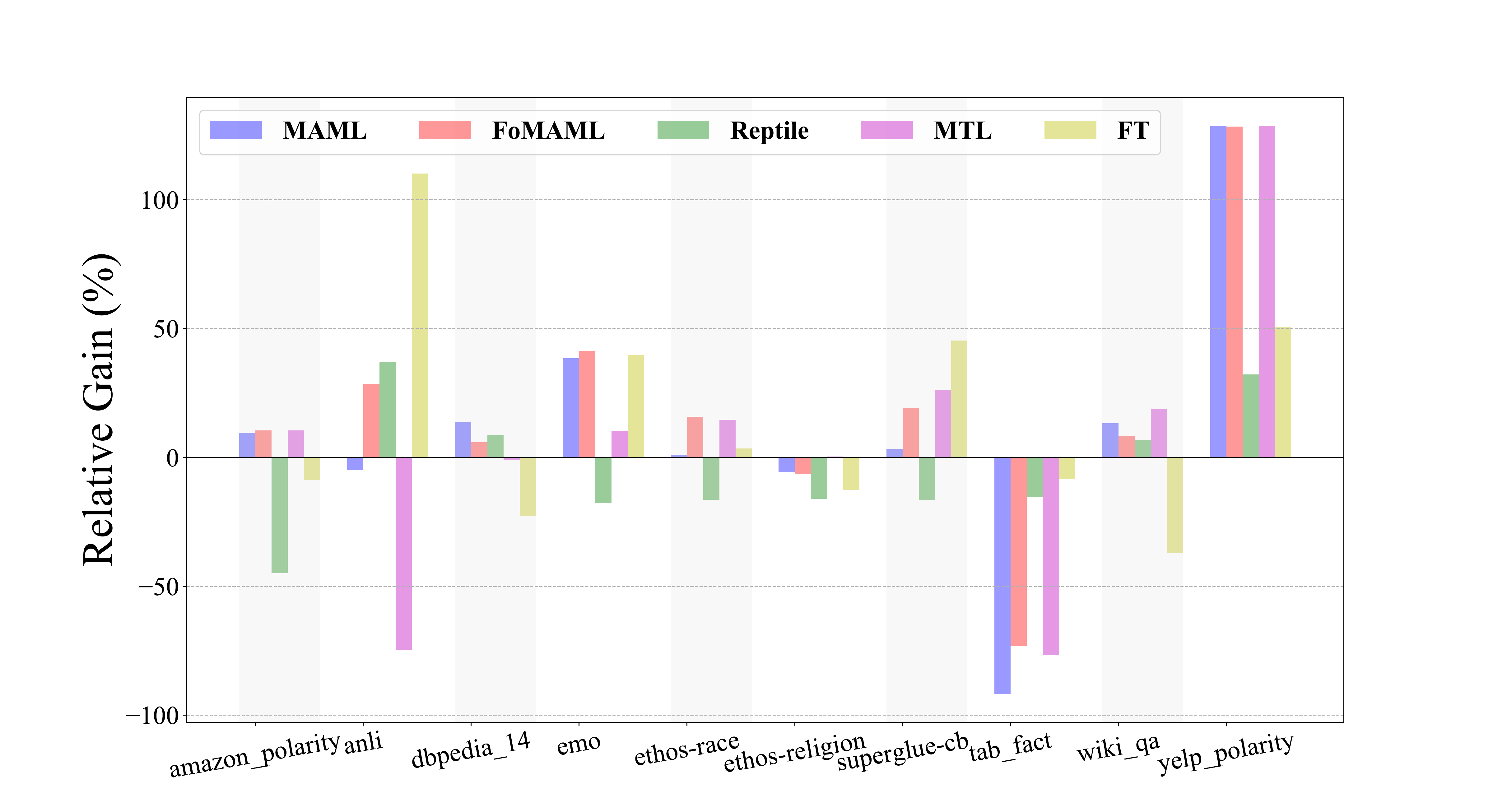}
    \caption{Both (Classification + Non-Classification) to Classification (Relative Gain)}
    \label{fig:both2cls}
\end{figure}

\begin{figure}[t]
    \centering
    \includegraphics[width=0.48\textwidth]{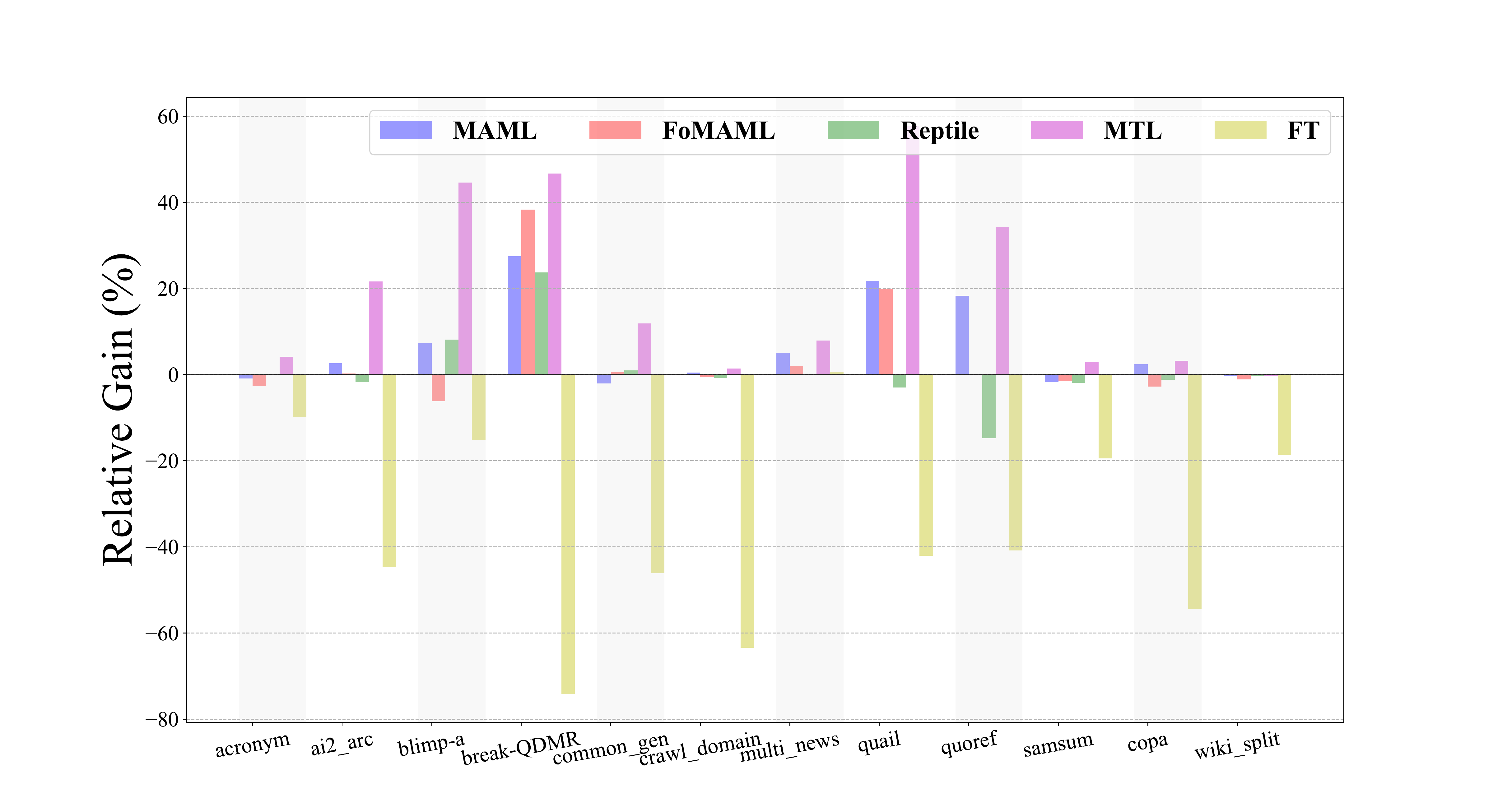}
    \caption{Non-Classification to Non-Classification (Relative Gain)}
    \label{fig:nocls2nocls} 
\end{figure}

\begin{figure}[t]
    \centering
    \includegraphics[width=0.48\textwidth]{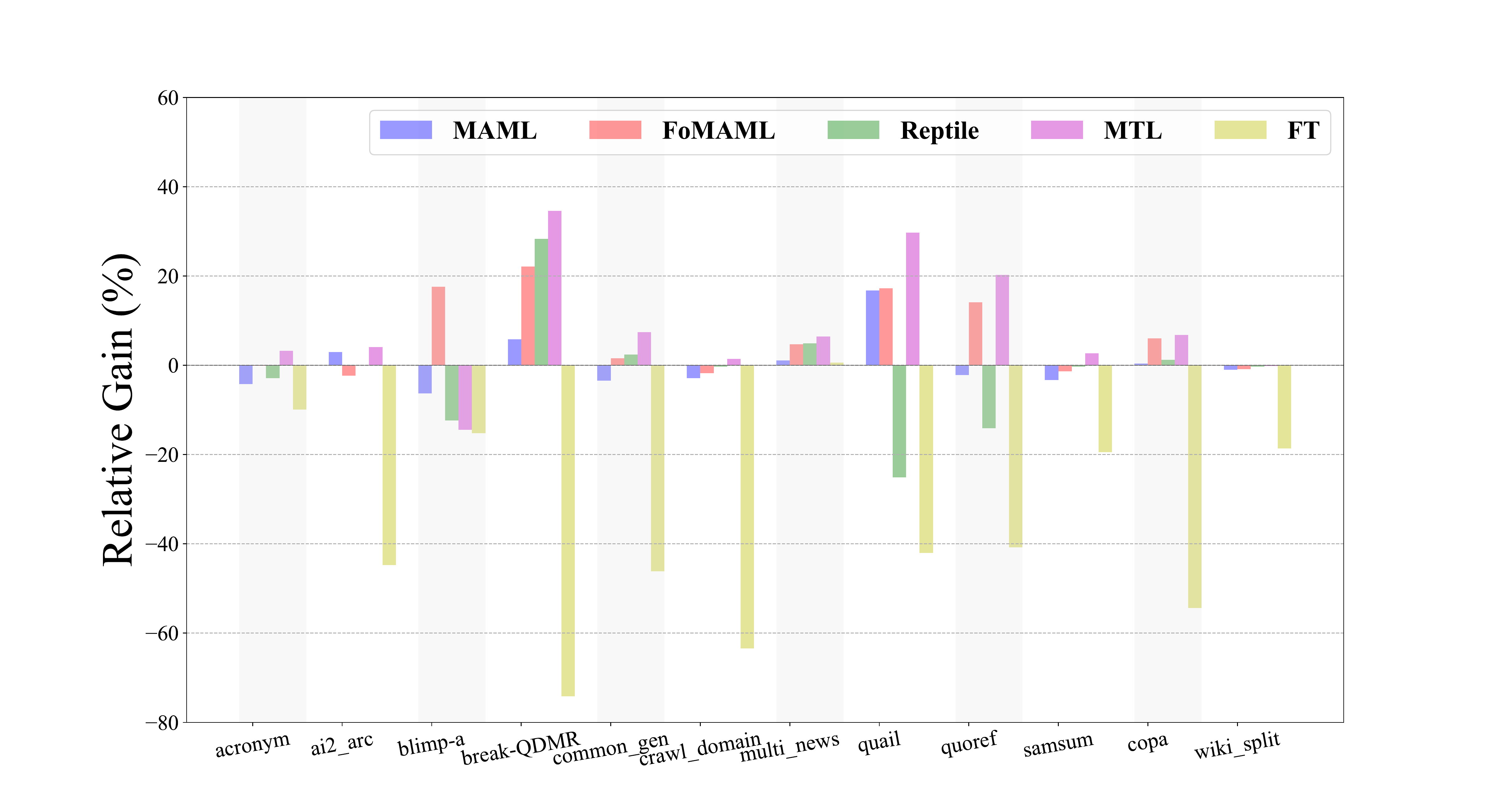}
    \caption{Classification to Non-Classification (Relative Gain)}
    \label{fig:cls2nocls}
\end{figure}

\begin{figure}[t]
    \centering
    \includegraphics[width=0.48\textwidth]{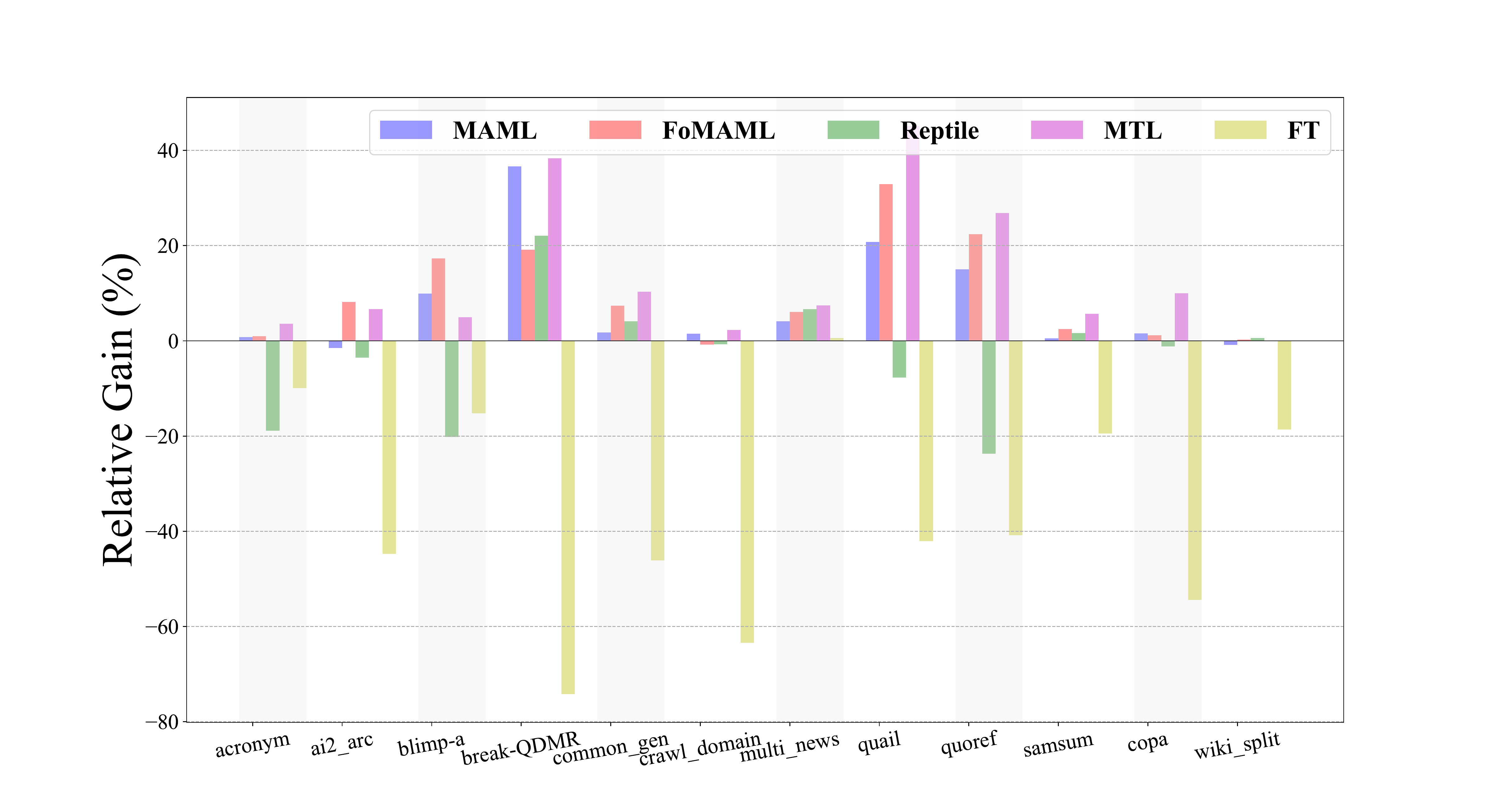}
    \caption{Both (Classification + Non-Classification) to Non-Classification (Relative Gain)}
    \label{fig:both2nocls}
\end{figure}

\begin{figure}[t]
    \centering
    \includegraphics[width=0.48\textwidth]{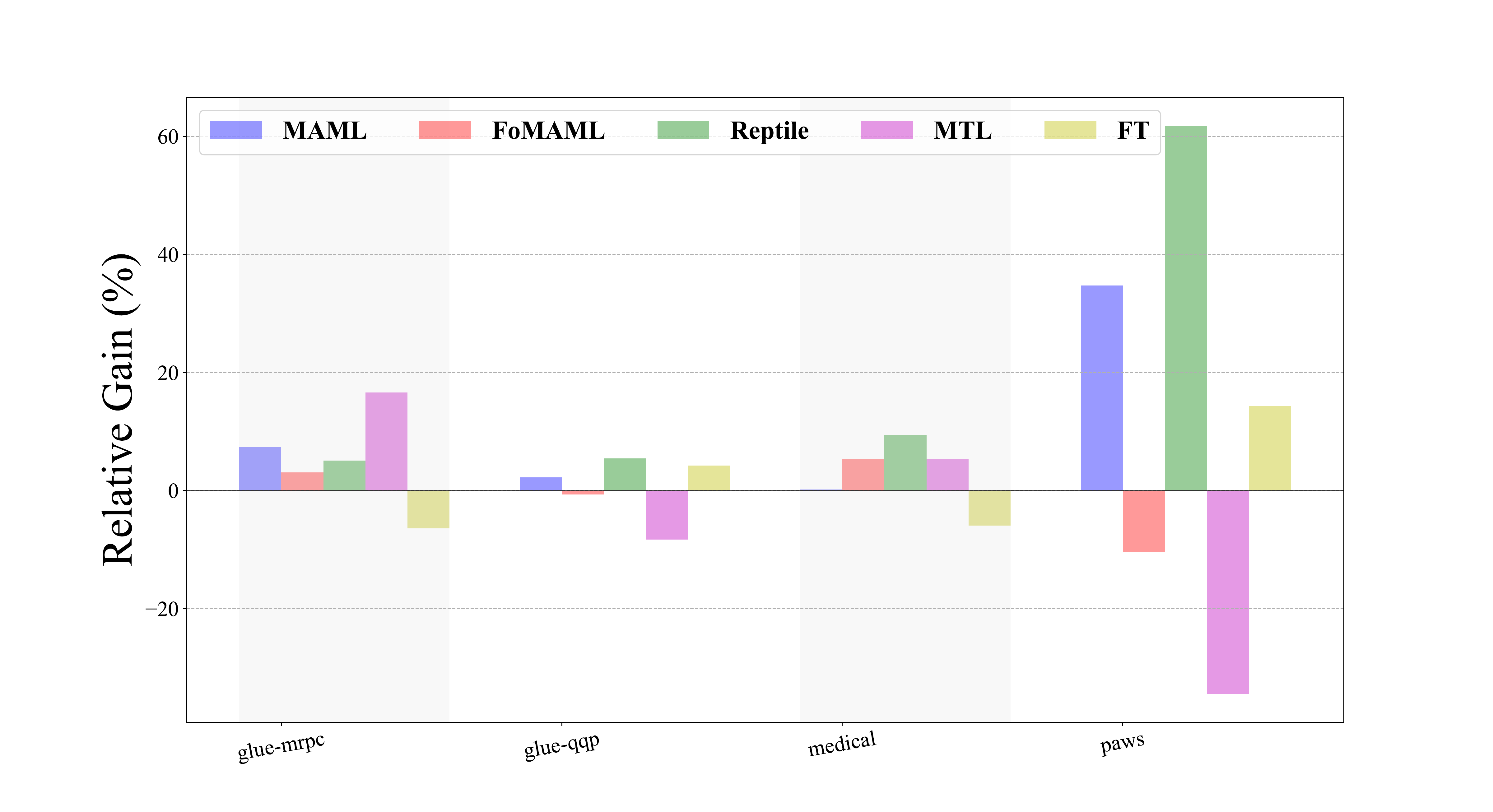}
    \caption{Non-Paraphrase Classification to Paraphrase (Relative Gain)}
    \label{fig:nopara2para}
\end{figure}

\begin{figure*}[t]
    \centering
    \includegraphics[width=1.00\textwidth]{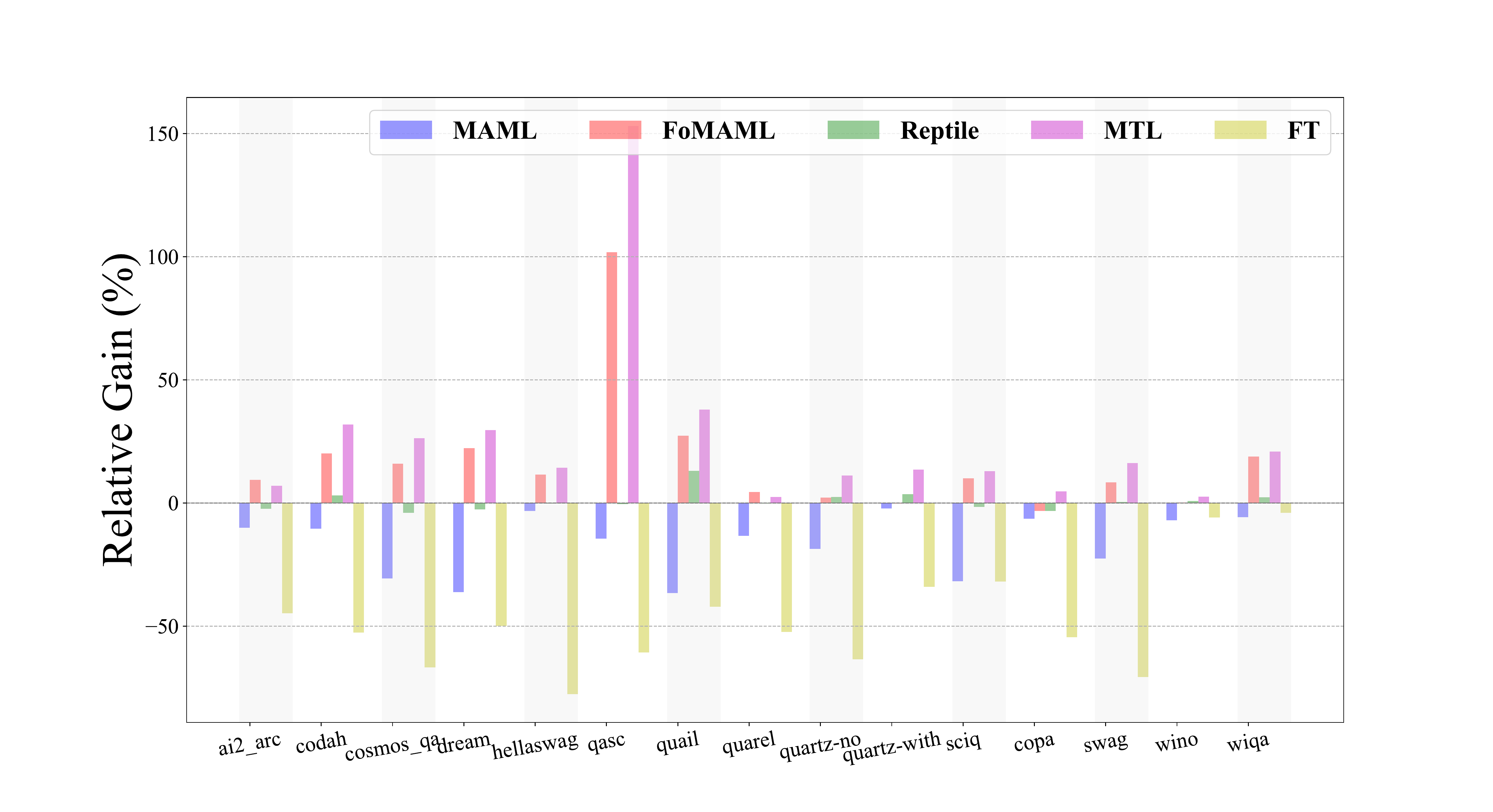}
    \caption{QA to QA (Relative Gain)}
    \label{fig:qa2qa}
\end{figure*}

\begin{figure}[t]
    \centering
    \includegraphics[width=0.48\textwidth]{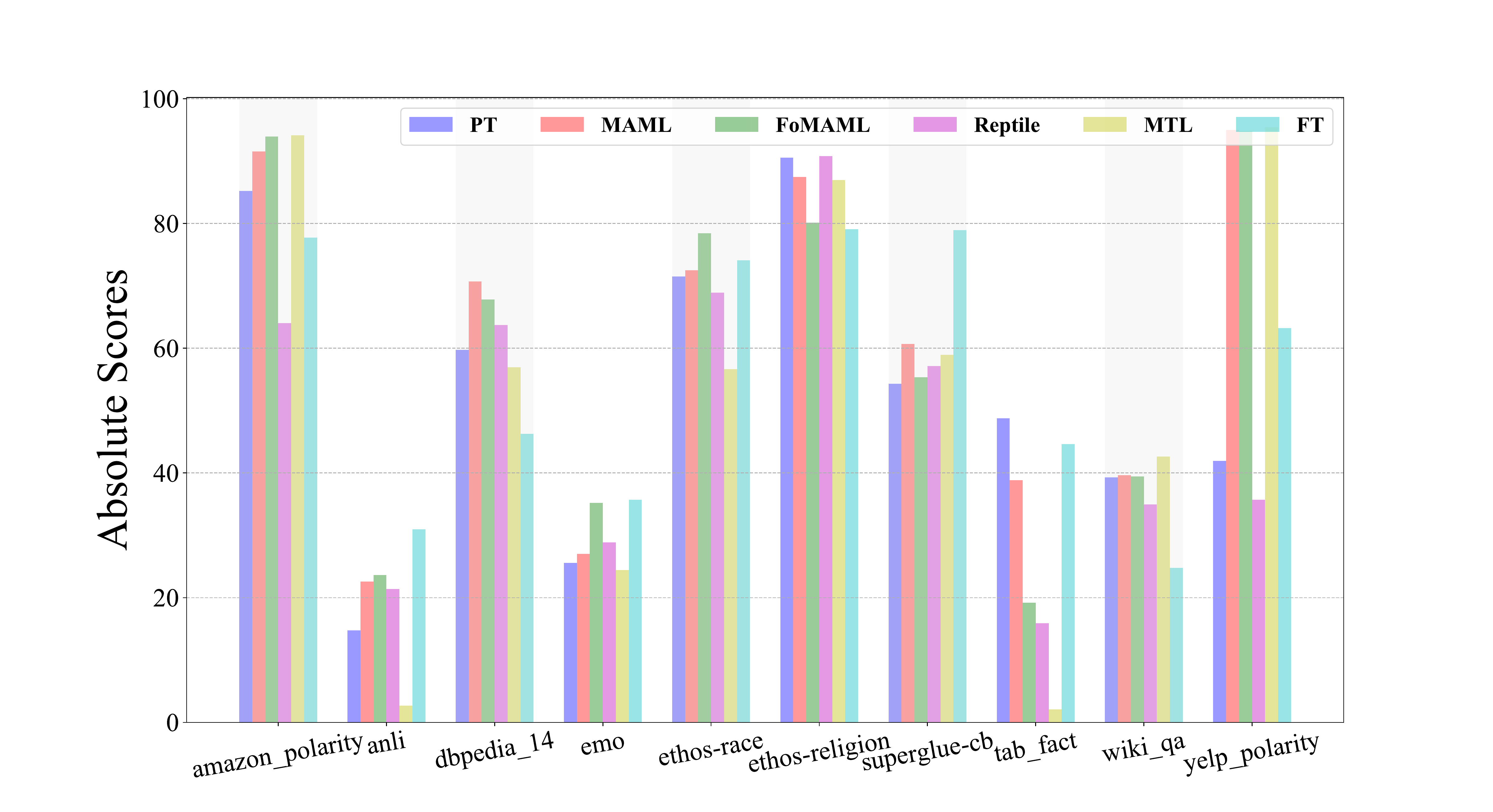}
    \caption{Classification to Classification (Absolute Scores)}
    \label{fig:cls2cls_absolute}
\end{figure}

\begin{figure}[t]
    \centering
    \includegraphics[width=0.48\textwidth]{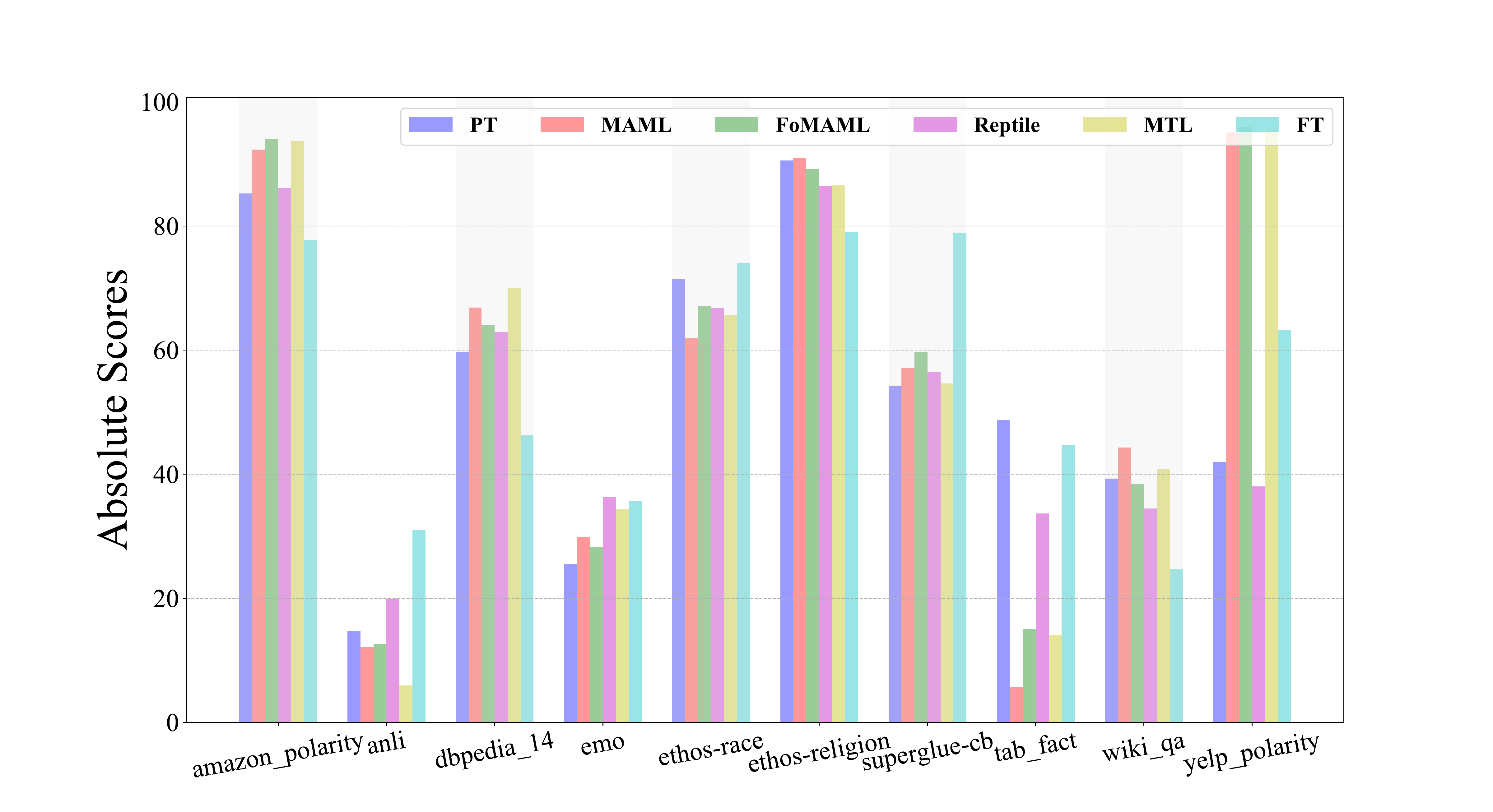}
    \caption{Non-Classification to Classification (Absolute Scores)}
    \label{fig:nocls2cls_absolute}
\end{figure}

\begin{figure}[t]
    \centering
    \includegraphics[width=0.48\textwidth]{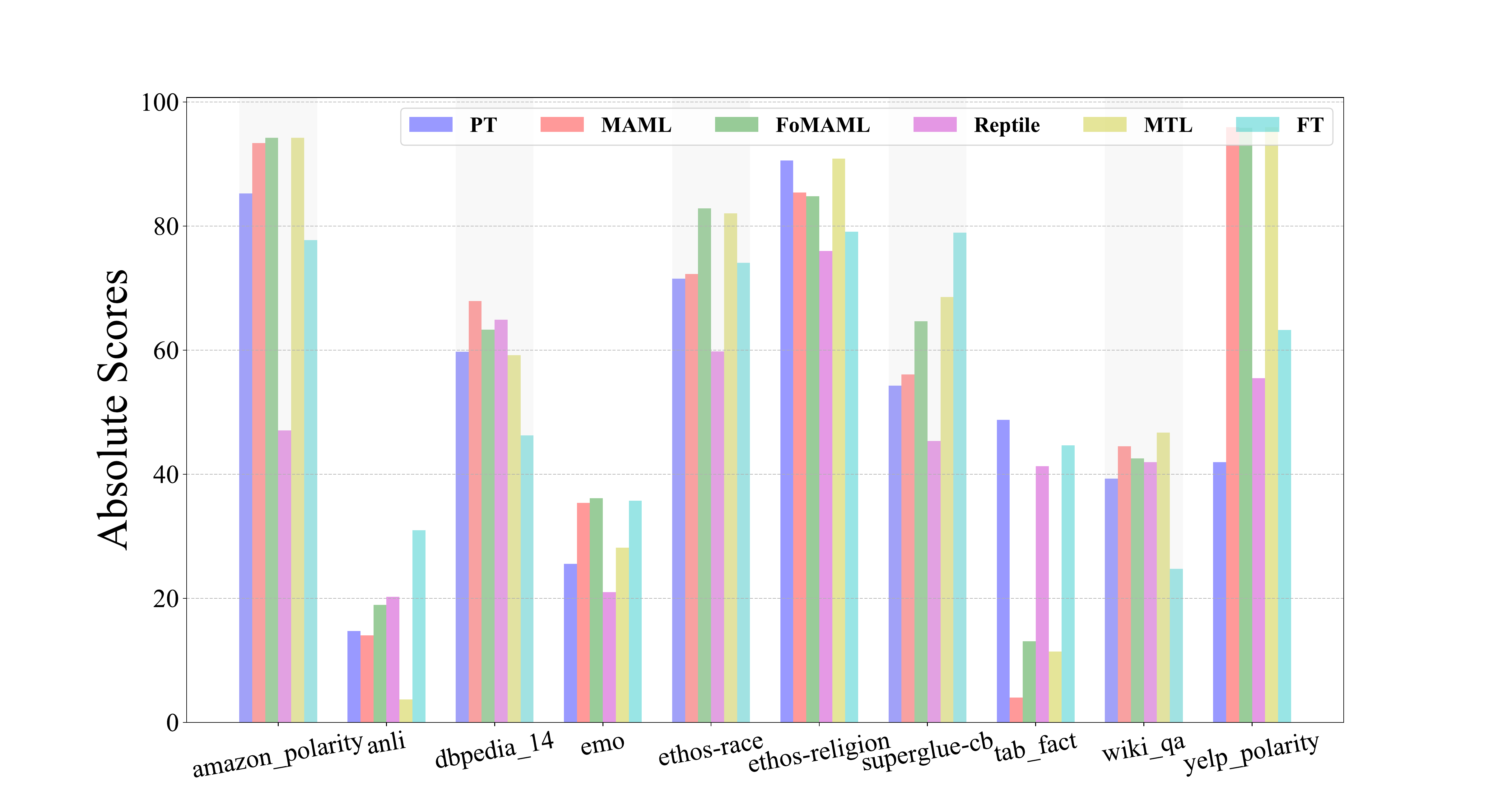}
    \caption{Both (Classification + Non-Classification) to Classification (Absolute Scores)}
    \label{fig:both2cls_absolute}
\end{figure}

\begin{figure}[t]
    \centering
    \includegraphics[width=0.48\textwidth]{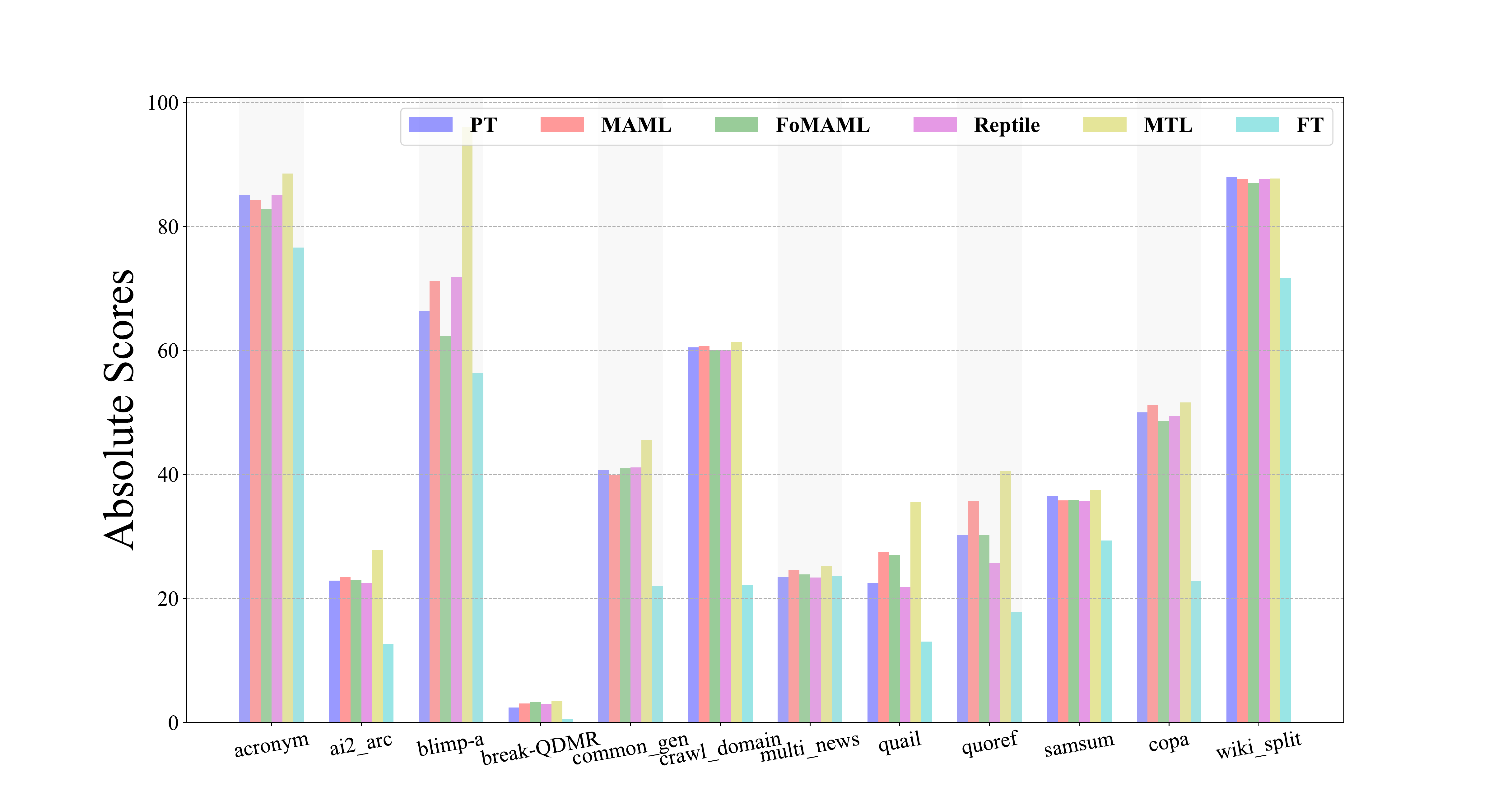}
    \caption{Non-Classification to Non-Classification (Absolute Scores)}
    \label{fig:nocls2nocls_absolute}
\end{figure}

\begin{figure}[t]
    \centering
    \includegraphics[width=0.48\textwidth]{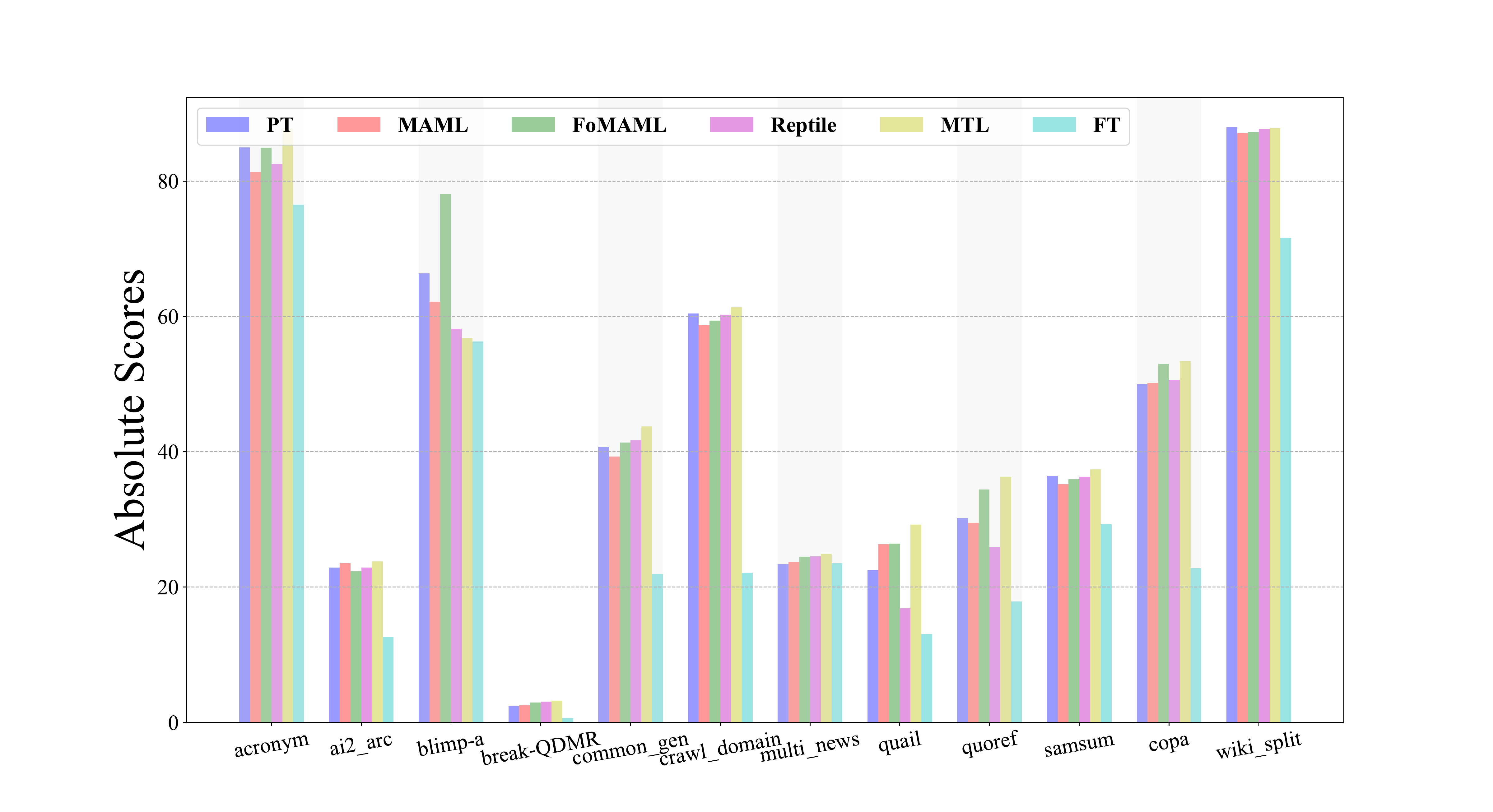}
    \caption{Classification to Non-Classification (Absolute Scores)}
    \label{fig:cls2nocls_absolute}
\end{figure}

\begin{figure}[t]
    \centering
    \includegraphics[width=0.48\textwidth]{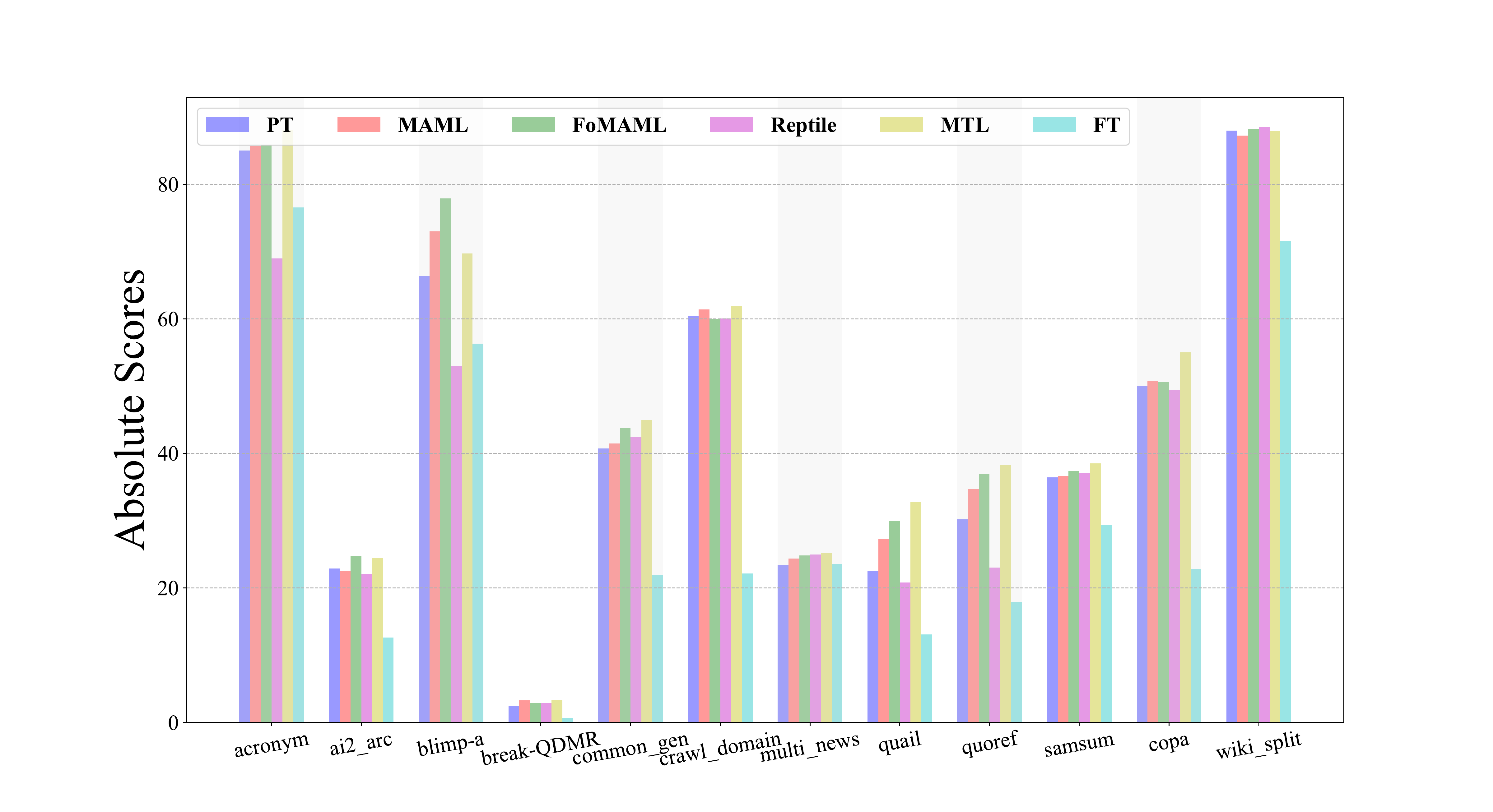}
    \caption{Both (Classification + Non-Classification) to Non-Classification (Absolute Scores)}
    \label{fig:both2nocls_absolute}
\end{figure}

\begin{figure}[t]
    \centering
    \includegraphics[width=0.48\textwidth]{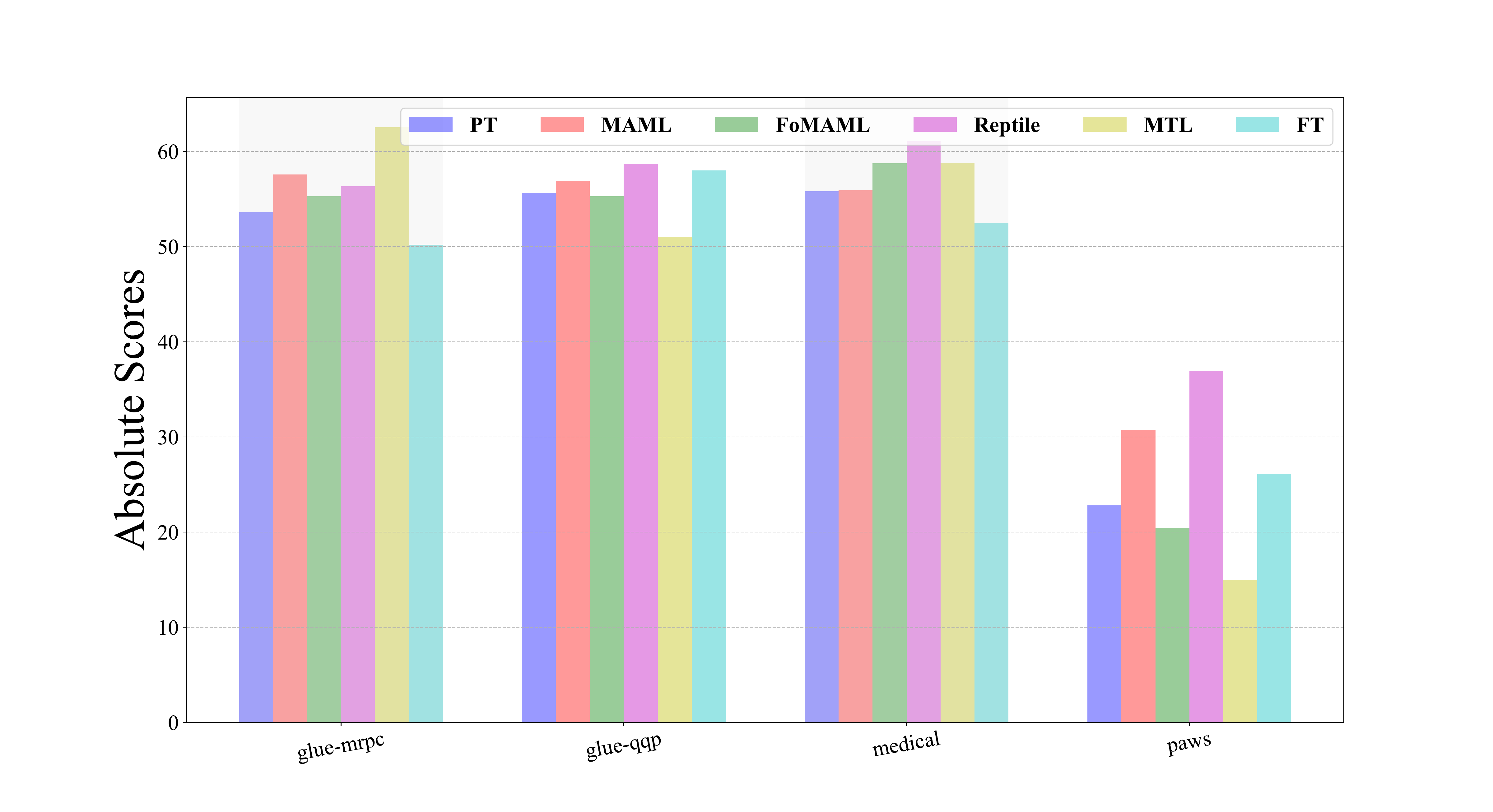}
    \caption{Non-Paraphrase Classification to Paraphrase (Absolute Scores)}
    \label{fig:nopara2para_absolute}
\end{figure}

\begin{figure*}[t]
    \centering
    \includegraphics[width=1.00\textwidth]{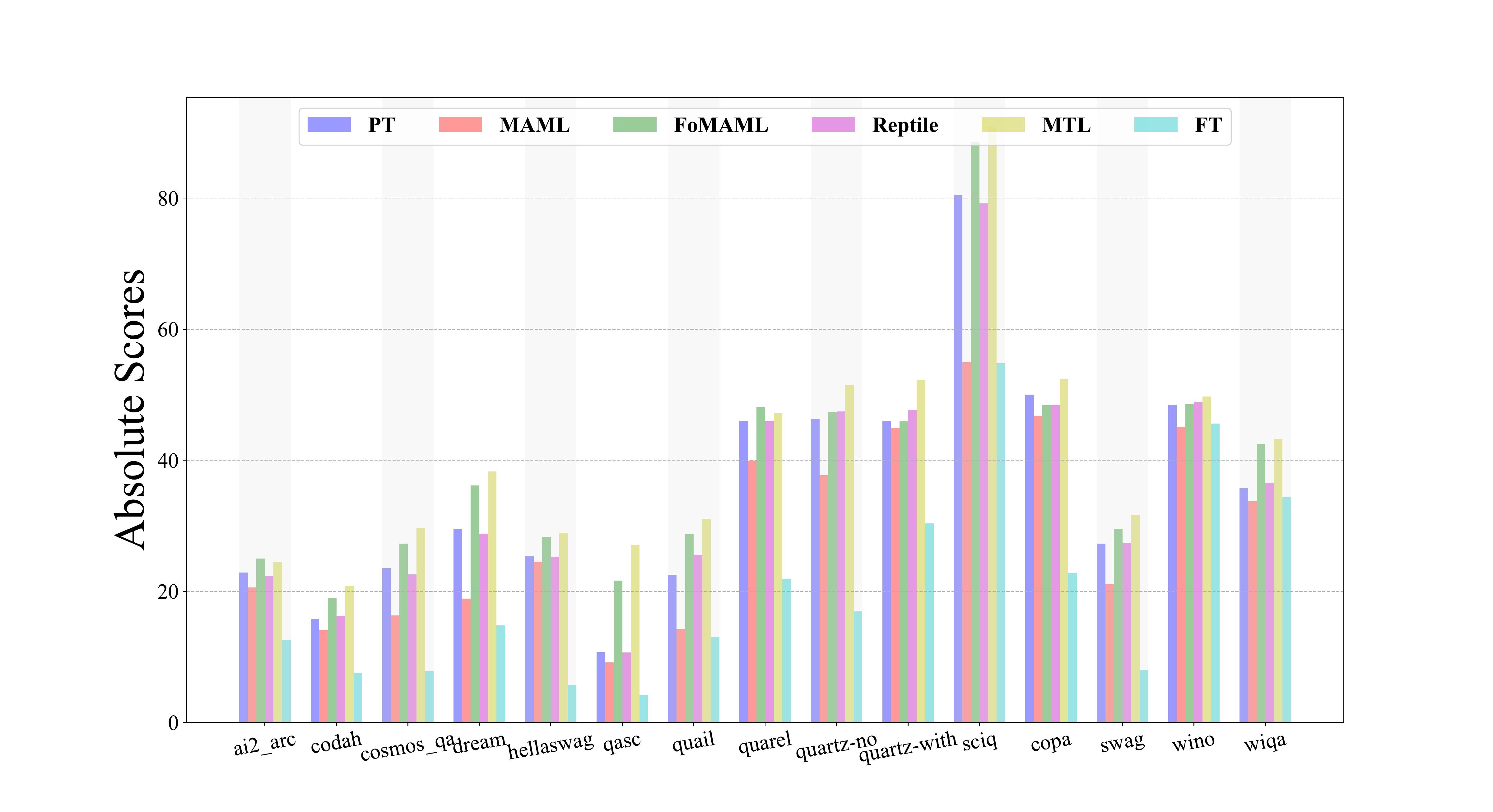}
    \caption{QA to QA (Absolute Scores)}
    \label{fig:qa2qa_absolute}
\end{figure*}

\end{document}